\documentclass[10pt]{article} 

\usepackage[preprint]{tmlr}

\usepackage{amsmath,amsfonts,bm}

\def\eqref#1{equation~\ref{#1}}

\def\1{\bm{1}}

\DeclareMathAlphabet{\mathsfit}{\encodingdefault}{\sfdefault}{m}{sl}
\SetMathAlphabet{\mathsfit}{bold}{\encodingdefault}{\sfdefault}{bx}{n}

\usepackage{multirow}
\usepackage{graphicx}
\usepackage{amsfonts}
\usepackage{amsmath}
\usepackage{amsthm}
\usepackage{algorithm}
\usepackage{algorithmic}
\usepackage[switch]{lineno}
\usepackage{booktabs}
\usepackage{nicefrac} 
\usepackage{xcolor}  
\usepackage{mathtools}
\usepackage{array}
\usepackage{hyperref}
\usepackage{url}

\title{Learning Energy-Based Generative Models via Potential Flow: A Variational Principle Approach to Probability Density Homotopy Matching}

\author{\name Junn Yong Loo \email loo.junnyong@monash.edu\\
\addr Monash University Malaysia
\AND
\name Michelle Adeline \email made0008@student.monash.edu\\
\addr Monash University Malaysia
\AND
\name Julia Kaiwen Lau \email julia.lau@monash.edu\\
\addr Monash University Malaysia
\AND
\name Fang Yu Leong \email leong.fangyu@monash.edu\\
\addr Monash University Malaysia
\AND
\name Hwa Hui Tew \email hwa.tew@monash.edu\\
\addr Monash University Malaysia
\AND
\name Arghya Pal \email arghya.pal@monash.edu\\
\addr Monash University Malaysia
\AND
\name Vishnu Monn Baskaran \email vishnu.monn@monash.edu\\
\addr Monash University Malaysia
\AND
\name Chee-Ming Ting \email ting.cheeming@monash.edu\\
\addr Monash University Malaysia
\AND
\name Rapha\"{e}l C.-W. Phan \email raphael.phan@monash.edu\\
\addr Monash University Malaysia
}

\def\month{05}  
\def\year{2025} 
\def\openreview{\url{https://openreview.net/forum?id=vc7poEYOFK}} 

\theoremstyle{plain}

\newtheorem{theorem}{Theorem}
\newtheorem{proposition}[theorem]{Proposition}

\theoremstyle{definition}

\theoremstyle{remark}
\newtheorem{remark}{Remark}
\DeclareMathOperator{\mean}{\mathbb{E}}

\DeclareMathOperator{\tr}{\mathrm{tr}}
\DeclareMathOperator{\cov}{\mathrm{Cov}}

\DeclareMathOperator{\KLD}{\mathcal{D}_\mathrm{KL}}
\DeclareMathOperator{\sgmd}{\mathrm{sigmoid}}

\begin{document}

\maketitle

\begin{abstract}
Energy-based models (EBMs) are a powerful class of probabilistic generative models due to their flexibility and interpretability. However, relationships between potential flows and explicit EBMs remain underexplored, while contrastive divergence training via implicit Markov chain Monte Carlo (MCMC) sampling is often unstable and expensive in high-dimensional settings. In this paper, we propose Variational Potential Flow Bayes (VPFB), a new energy-based generative framework that eliminates the need for implicit MCMC sampling and does not rely on auxiliary networks or cooperative training. VPFB learns an energy-parameterized potential flow by constructing a flow-driven density homotopy that is matched to the data distribution through a variational loss minimizing the Kullback-Leibler divergence between the flow-driven and marginal homotopies. This principled formulation enables robust and efficient generative modeling while preserving the interpretability of EBMs. Experimental results on image generation, interpolation, out-of-distribution detection, and compositional generation confirm the effectiveness of VPFB, showing that our method performs competitively with existing approaches in terms of sample quality and versatility across diverse generative modeling tasks.
\end{abstract}

\section{Introduction}

Energy-based models (EBMs) have emerged 
as a flexible and expressive class of probabilistic generative models \citep{Shortrun_MCMC,IGEBM,LSD,Flow_Contrastive,IGEBMpp,Diffusion_Recovery,JEM,SADA_JEM,Cooperative_Diffusion_Recovery}. 
By assigning a potential energy that correlates with the unnormalized data likelihood \citep{Howto_EBM}, EBMs offer a structured energy landscape for probability density estimation, providing several notable advantages. 
First, EBMs are interpretable, as the underlying energy function can be visualized in terms of energy surfaces. Second, they are highly expressive and do not impose strong architectural constraints \citep{BondTaylor}, enabling them to capture complex data distributions. Third, EBMs exhibit inherent robustness to Out-of-Distribution (OOD) inputs, given that regions with low likelihood are naturally penalized \citep{IGEBM,JEM}. 
Building on their origins in Boltzmann machines \citep{Contrastive_Divergence}, EBMs also share conceptual ties with statistical physics, allowing practitioners to adapt physical insights and tools for model design and analysis \citep{Feinauer}. They have demonstrated promising performance in various applications beyond image modeling, including text generation \citep{ResidualEBM}, robot learning \citep{Planning}, point cloud synthesis \citep{GPointNet},  trajectory prediction \citep{LB-EBM,SEEM}, molecular design \citep{GraphEBM}, and anomaly detection \citep{Yoon}.

Despite these advantages, training deep EBMs often relies on implicit Markov Chain Monte Carlo (MCMC) sampling for contrastive divergence. In high-dimensional settings, MCMC suffers from poor mode mixing and slow mixing \citep{IGEBM,Shortrun_MCMC,Flow_Contrastive,JEM,MCMC_ShouldMix,BondTaylor}, yielding biased estimates that may optimize unintended objectives \citep{LSD}. Truncated chains, in particular, can lead models to learn an implicit sampler rather than a true density, which prevents valid steady-state convergence and inflates computational overhead. As a result, the generated samples can deviate significantly from the target distribution \citep{LSD}. To mitigate these issues, some works propose auxiliary or cooperative strategies that learn complementary models to either avoid MCMC via variational inference \citep{VAEBM} or combine short-run MCMC refinements with learned generator distributions \citep{CoopNets,VERA,HatEBM}. Nevertheless, these approaches could complicate model architectures and training procedures.

In parallel, flow-based models have advanced generative modeling by leveraging continuous normalizing flows and optimal transport techniques to surpass diffusion models in sample quality and efficiency \citep{DDPMpp,NCSNpp}. Notable examples include Flow Matching \citep{FlowMatching}, which models diffeomorphic mappings between noise and data; Rectified Flow \citep{RectifiedFlow}, which optimizes sampling paths; Stochastic Interpolants \citep{StochasticInterpolant,Rezende,NeuralODE}, which incorporate stochastic processes into flows for complex data geometries; Schrödinger Bridge Matching \citep{BridgeMatching}, which integrates entropy-regularized optimal transport with diffusion; and Poisson Flow Generative Model (PFGM) \citep{PFGM}, which introduces an augmented space governed by the Poisson equation. However, these methods do not directly parameterize probability density and lack the theoretical advantages of EBMs, such as generating conservative vector fields aligned with log-likelihood gradients \citep{EnergyorScore}. 

Recent approaches, such as Action Matching \citep{ActionMatching}, explicitly model the energy (action) to generate data-recovery vector fields, thus providing a structured approach to learning conservative dynamics. Meanwhile, Diffusion Recovery Likelihood (DRL) \citep{Diffusion_Recovery} and Denoising Diffusion Adversarial EBMs (DDAEBM) \citep{DDAEBM} refine conditional EBMs by improving sampling efficiency and training stability through diffusion-based probability paths. However, a direct connection between energy-parameterized flow models and explicit (marginal) EBMs remains unexplored, limiting the application of flow-based techniques for learning EBMs. Furthermore, existing generative models have yet to adopt variational formulations, such as the Deep Ritz method, to align the evolution of density paths.

To address the computational challenges of existing energy-based methods, we propose Variational Potential Flow Bayes (VPFB), a novel generative framework grounded in variational principles that eliminates the need for auxiliary models and implicit MCMC sampling. VPFB employs the Deep Ritz method to learn an energy-parameterized potential flow, ensuring alignment between the flow-driven density homotopy and the data-recovery likelihood homotopy. To address the intractability of homotopy matching, we formulate a variational loss function that minimizes the Kullback-Leibler (KL) divergence between these density homotopies. Additionally, we validate the learned potential energy as an effective parameterization of the stationary Boltzmann energy. Through empirical validations, we benchmark VPFB against state-of-the-art generative models, showcasing its competitive performance in Fréchet Inception Distance (FID) for image generation and excellent OOD detection with high Area Under the Receiver Operating Characteristic Curve (AUROC) scores across multiple datasets.

\section{Background and Related Works}

In this section, we provide an overview of EBMs, particle flow, and the Deep Ritz method, collectively forming the cornerstone of the proposed VPFB framework. 

\subsection{Energy-based Models (EBMs)}

Denote $\bar{x} \in \Omega \subseteq \mathbb{R}^n$ as the training data, EBMs approximate the data likelihood ${p}_{\text{data}}(\bar{x})$ via defining a Boltzmann distribution
\begin{align} \label{eq:Boltzmann_distribution}
\begin{split}
{p}_{B}(x) = \frac{e^{\Phi_{B}(x)}}{Z}
\end{split}
\end{align}
where $\Phi_{B}$ is the Boltzmann energy parameterized via neural networks and $Z = {\int_{\Omega} e^{\Phi_{B}(x)} \, dx}$ is the normalizing constant.
Given that this partition function is analytically intractable for high-dimensional data, EBMs perform the Maximum Likelihood Estimation (MLE) by minimizing the negative log-likelihood loss $\mathcal{L}_{\mathrm{MLE}}(\theta) = - \mean_{{p}_{\text{data}}(\bar{x})} [\log {p}_{B}(\bar{x})] = \mean_{{p}_{\text{data}}(\bar{x})} \big[ \Phi_{B}(\bar{x}) \big] - \mean_{{p}_{\text{data}}(\bar{x})} \big[ \log Z \big]$. The gradient of this MLE loss with respect to model parameters $\theta$ is approximated via the contrastive divergence \citep{Contrastive_Divergence} loss
$\nabla_{\theta} \mathcal{L}_{\mathrm{MLE}} = \mean_{{p}_{\text{data}}(\bar{x})} \big[ \nabla_{\theta} \Phi_{B}(\bar{x}) \big] - \mean_{{p}_{B}(x)} \big[ \nabla_{\theta} \Phi_{B}(x) \big]$.
Nonetheless, EBMs are computationally intensive due to the implicit MCMC generating procedure required for generating negative samples $x \sim {p}_{B}(x)$ implicitly during training.



\subsection{Particle Flow}

Particle flow, introduced by \citet{Daum}, is a class of nonlinear Bayesian filtering (sequential inference) methods designed to approximate the posterior distribution $p(x_{t} \mid \bar{x}_{\leq t})$ of the sampling process given observations. While closely related to normalizing flows \citep{Rezende} and neural ordinary differential equations (ODEs) \citep{NeuralODE}, these frameworks do not explicitly accommodate a Bayes update.
Instead, particle flow achieves Bayes update $p(x_{t} \mid \bar{x}_{\leq t}) \propto p(x_{t} \mid \bar{x}_{<t}) \, p(\bar{x}_{t} \mid x_{t},\bar{x}_{<t})$ by transporting the prior samples $x_{t} \sim p(x_{t} \mid \bar{x}_{<t})$ through an ODE $\frac{dx}{dt} = v(x,t)$ parameterized by a velocity field $v(x,t)$, over pseudo-time $t \in [0, 1]$. The velocity field is designed such that the sample density follows a log-homotopy that induces the Bayes update. 
Despite its effectiveness in time-series inference \citep{Pal, PFBR, Yang_3} and its robustness against the curse of dimensionality \citep{CoD}, particle flow, particularly potential flow where the velocity field $v(x,t) = \Phi(x, t)$ is the gradient of potential energy, remains largely unexplored in energy-based generative modeling.


\subsection{Deep Ritz Method}
The Deep Ritz method is a deep learning-based variational numerical approach, originally proposed by \citet{Deep_Ritz}, for solving scalar elliptic partial differential equations (PDEs) in high dimensions.
Consider the following Poisson's equation, fundamental to many physical models:
\begin{align}
\begin{split} \label{eq:Dirichlet_problem}
\Delta_x u(x) = \Gamma(x), &\quad x \in \Omega \\
u(x) = 0, &\quad x \in \partial\Omega
\end{split}
\end{align}
where $\Delta_x$ is the Laplace operator, and $\partial \Omega$ denotes the boundary of $\Omega$. For a Sobolev function $u \in \mathcal{H}^1_0(\Omega)$ (definition in Proposition \ref{thm:proposition_2}) and square-integrable $\Gamma \in L^2(\Omega)$, the variational principle ensures that a weak solution of the Euler-Lagrange boundary value equation (\ref{eq:Dirichlet_problem}) is equivalent to the variational problem of minimizing the Dirichlet energy \citep{Deep_Ritz_revisited}, as follows:
\begin{align}
\begin{split} \label{eq:Dirichlet_energy}
u = \arg \underset{v}{\min} \int_{\Omega} \bigg( \, \frac{1}{2} \, \| \nabla_{x} v(x) \|^2 - \Gamma(x) \, v(x) \bigg) \, dx + \eta \int_{\partial \Omega} v(x)^2 \ dx
\end{split}
\end{align}
where $\nabla_{x}$ denotes the Del operator (gradient). In particular, the Deep Ritz method parameterizes the trial function $v$ using neural networks and performs the optimization (\ref{eq:Dirichlet_energy}) via stochastic gradient descent. 
To enforce the Dirichlet boundary condition, the second component of the Dirichlet energy (\ref{eq:Dirichlet_energy}), weighted by a positive constant $\eta$, must be evaluated on the boundary $\partial \Omega$. This necessitates acquiring additional boundary samples $x \in \partial \Omega$ during neural network training, thereby introducing extra computational overhead.
The Deep Ritz method is 
predominantly applied for finite element analysis \citep{Deep_Ritz_FEM} due to its versatility and effectiveness in handling high-dimensional PDE systems. In \citep{Olmez}, the Deep Ritz method is employed to solve the density-weighted Poisson equation arising from the feedback particle filter \citep{Yang_1}. However, its application in generative modeling remains unexplored.


\section{Variational Potential Flow Bayes (VPFB)}
In this section, we introduce VPFB, a novel generative modeling framework
inspired by particle flow and the Deep Ritz method. VPFB encompasses four key elements: constructing a Bayesian marginal homotopy between the Gaussian prior and data likelihood (Section \ref{ssect:LogHomotopy}), designing a potential flow that aligns the flow-driven homotopy with the marginal homotopy (Section \ref{modeling_potential_flow}), formulating a variational loss function using the Deep Ritz method (Section \ref{sec: variational_loss}), and establishing connections between homotopy matching, diffusion, and EBMs (Section \ref{sec: connections_diffusion_ebm}).

\subsection{Interpolating Between Prior and Data Likelihood: Log-Homotopy Bayesian Transport} \label{ssect:LogHomotopy}
Let $\bar{x} \in \Omega$ denote the training data, with likelihood $p_{\text{data}}(\bar{x})$, 
and let $x \in \Omega$ represent the generative samples. First, we define a Gaussian prior $q(x) = \mathcal{N}(0, \omega^2 I)$ and a Gaussian conditional data likelihood $p(\bar{x} \mid x) = \mathcal{N}(\bar{x}; x, \nu^2 I)$, both with isotropic covariances. This data likelihood satisfies the state space model $x = \bar{x} + \nu \, \epsilon$, where $\epsilon$ is the standard Gaussian noise. The standard deviation $\nu$ is usually set to be small so that $x$ closely resembles $\bar{x}$. The aim of flow-based generative modeling is to learn a density homotopy (path) interpolating between the prior and the data likelihood for generative modeling.
On that account, consider the following conditional (data-conditioned) probability density log-homotopy $\rho: \Omega^{2} \times [0, 1] \rightarrow \mathbb{R}$:
\begin{align} \label{eq:unmarginalized_homotopy_h}
\begin{split}
{\rho}(x \mid \bar{x},t) = \frac{e^{h(x \mid \bar{x},t)}}{\int_{\Omega} e^{h(x \mid \bar{x},t)} \, dx}
\end{split}
\end{align}
where $h: \Omega^{2} \times [0, 1] \rightarrow \mathbb{R}$ is a log-linear function:
\begin{align} \label{eq:log-homotopy_f}
\begin{split}
h(x \mid \bar{x}, t) = \alpha(t) \, \log q(x) + \beta(t) \, \log p(\bar{x} \mid x)
\end{split}
\end{align}
where $\alpha: [0, 1] \rightarrow [0, 1]$ and $\beta: [0, 1] \rightarrow [0, 1]$ are both monotonically increasing functions parameterized by time $t$. The following proposition shows that this log-homotopy transformation results in a Gaussian perturbation kernel.

\begin{proposition}\label{thm:proposition_1}
Consider a Gaussian prior $q(x) = \mathcal{N}(x; 0, \omega^2 I)$ and a conditional data likelihood $p(\bar{x} \mid x) = \mathcal{N}(\bar{x}; x, \nu^2 I)$.
The log-homotopy transport (\ref{eq:unmarginalized_homotopy_h}) corresponds to a Gaussian perturbation kernel ${\rho}(x \mid \bar{x},t) = \mathcal{N}\big(x; \mu(t) \, \bar{x}, \sigma(t)^2 I\big)$, 
characterized by the time-varying mean and standard deviation:
\begin{align} \label{eq:pertubation_kernel}
\begin{split} 
\mu(t) &= \sgmd \Bigg( \log \bigg( \frac{\beta(t)}{\alpha(t)} \, \frac{\omega^2}{\nu^2} \bigg) \Bigg), \quad
\sigma(t) 
= \sqrt{\frac{\nu^2}{\beta(t)} \, \mu(t)}
\end{split}
\end{align}
where $\sgmd(z) = \frac{1}{1+e^{-z}}$ denotes the logistic (sigmoid) function. 
\end{proposition}

\begin{proof}
Refer to Appendix \ref{Appendix:A}.
\end{proof}

Hence, the density homotopy \eqref{eq:unmarginalized_homotopy_h} represents a tempered Bayesian transport mapping from the Gaussian prior $q(x)$ to the posterior kernel
\begin{align} \label{eq:posterior_kernel}
\begin{split}
{\rho}(x \mid \bar{x}, 1) 
&= \frac{e^{h(x \mid \bar{x},1)}}{\int_{\Omega} e^{h(x \mid \bar{x},1)} \, dx} 
= \frac{p(\bar{x} \mid x) \, q(x)}{\int_{\Omega} p(\bar{x} \mid x) \, q(x) \, dx}
= p(x \mid \bar{x})
\end{split}
\end{align}
which is the maximum a posteriori estimation centered on discrete data samples.
To approximate the intractable data likelihood, we can then consider the following marginal probability density homotopy:
\begin{align} \label{eq:marginalized_homotopy_h}
\begin{split}
\bar{\rho}(x,t) = \int_{\Omega} {p}_{\text{data}}(\bar{x}) \, {\rho}(x \mid \bar{x},t) \; d\bar{x},
\end{split}
\end{align}
where it remains that $p(x,0) = q(x)$, and we have $\bar{\rho}(x,1) = \int_{\Omega} {p}_{\text{data}}(\bar{x}) \, p(x \mid \bar{x}) \, d\bar{x} = p(x)$. 
Therefore, this marginal homotopy defines a data-recovery path interpolation between the Gaussian prior $q(x)$ and the approximate data likelihood $p(x)$. In particular, $p(x)$ represents a Bayesian approximation of the true data likelihood, by convolving the discrete data likelihood ${p}_{\text{data}}(\bar{x})$ with the posterior distribution $p(x \mid \bar{x})$. 
Nevertheless, the marginalization in (\ref{eq:marginalized_homotopy_h}) is intractable, thereby precluding a closed-form solution for the marginal homotopy. To overcome this challenge, we propose a potential flow-driven density homotopy, whose time evolution is aligned with this data-recovery marginal homotopy.


\subsection{Modeling Potential Flow in a Data-Recovery Homotopy Landscape}\label{modeling_potential_flow}
Our goal is to model a potential flow whose density evolution aligns with the marginal homotopy, thereby directing samples toward the data likelihood. We begin by deriving the time evolution of the marginal homotopy in the following proposition.

\begin{proposition} \label{thm:proposition_2} 
Consider the conditional homotopy ${\rho}(x \mid \bar{x},t)$ in (\ref{eq:unmarginalized_homotopy_h}) with Gaussian conditional data likelihood $p(\bar{x} \mid x) = \mathcal{N}(\bar{x}; x, \nu^2 I)$. Then, the time evolution (derivative) of the marginal homotopy $\bar{\rho}(x,t)$ is given by the following partial differential equation (PDE):

\begin{align}
\begin{split} \label{eq:homotopy_PDE}
\frac{\partial \bar{\rho}(x,t)}{\partial t} = - \frac{1}{2} \, \mean_{{p}_{\text{data}}(\bar{x})} \Big[ {\rho}(x \mid \bar{x},t) \, \Big( \gamma(x, \bar{x}, t) - \bar{\gamma}(x, \bar{x}, t) \Big) \Big]
\end{split}
\end{align}
where $\gamma$ denotes the innovation term
\begin{align}
\begin{split} \label{eq:innovation}
\gamma(x, \bar{x}, t) = \frac{\dot{\alpha}(t)}{\omega^2} \, \|x\|^2 + \frac{\dot{\beta}(t)}{\nu^2} \, \|x - \bar{x}\|^2
\end{split}
\end{align}
Here, $\dot{\alpha}(t)$ and $\dot{\beta}(t)$ denote the time-derivatives, and $\bar{\gamma}(x, \bar{x}, t) = \mean_{{\rho}(x \mid \bar{x},t)} [\gamma(x, \bar{x}, t)]$ denotes the expectation.
\end{proposition}

\begin{proof}
Refer to Appendix \ref{Appendix:B}.
\end{proof}

A potential flow involves subjecting the prior samples to an energy-generated velocity field, where their trajectories ($x(t)$) satisfy the following ODE:
\begin{align} \label{eq:particle_flow_ODE}
\frac{dx(t)}{dt} = \nabla_{x} \Phi(x, t)
\end{align}
where $\Phi: \Omega \times [0, 1] \rightarrow \mathbb{R}$ is a scalar potential energy, and $\nabla_{x}$ denotes the Del operator (gradient) with respect to the data samples $x(t)$. The vector field $\nabla_{x} \Phi \in \Omega$ represents the divergence (irrotational) component in the Helmholtz decomposition.
By incorporating this potential flow, the flow-driven density homotopy ${\rho}_{\Phi}(x,t)$ evolves via the continuity equation \citep{gardiner}:
\begin{align}
\begin{split} \label{eq:Kolmogorov_forward}
\frac{\partial {\rho}_{\Phi}(x,t)}{\partial t} 
= - \, \nabla_{x} \cdot \Big( {\rho}_{\Phi}(x,t) \, \nabla_{x} \Phi(x,t) \Big)
\end{split}
\end{align}
which corresponds to the transport equation for modeling fluid advection. 
Our aim is to model the potential energy such that the evolution of the prior density under the potential flow emulates the evolution of the marginal homotopy. 
In other words, we seek to achieve homotopy matching, ${{\rho}_{\Phi}} \equiv \bar{\rho}$, by aligning their respective time evolutions as described in (\ref{eq:homotopy_PDE}) and (\ref{eq:Kolmogorov_forward}). This leads to the following PDE, which takes the form of a density-weighted Poisson equation:
\begin{align}
\begin{split} \label{eq:unmarginalized_PDE_intractable}
& \nabla_{x} \cdot \Big( {\rho}_{\Phi}(x,t) \, \nabla_{x} \Phi(x,t) \Big) 
= \frac{1}{2} \, \mean_{{p}_{\text{data}}(\bar{x})} \Big[ {\rho}(x \mid \bar{x},t) \, \Big( \gamma(x, \bar{x}, t) - \bar{\gamma}(x, \bar{x}, t) \Big) \Big]
\end{split}
\end{align}
However, this Poisson equation remains intractable due to the lack of a closed-form expression for ${\rho}_{\Phi}$. To overcome this limitation, we substitute the intractable ${\rho}_{\Phi}$ with the target marginal homotopy $\bar{\rho}$, enabling direct sampling and a variational principle approach. In the following proposition, we demonstrate that the revised Poisson's equation minimizes the KL divergence between the flow-driven and conditional homotopies, yielding statistically optimal homotopy matching.

\begin{proposition} \label{thm:proposition_3}
Consider a potential flow of the form (\ref{eq:particle_flow_ODE}) and given that $\Phi \in \mathcal{H}^1_0(\Omega, p)$, where $\mathcal{H}^n_0$ denotes the (Sobolev) space of $n$-times differentiable functions that are compactly supported, and square-integrable with respect to marginal homotopy $\bar{\rho}(x,t)$. 
Solving for the potential energy $\Phi(x)$ that satisfies the following density-weighted Poisson's equation:
\begin{align}
\begin{split} \label{eq:unmarginalized_PDE_tractable}
& \nabla_{x} \cdot \Big( \bar{\rho}(x,t) \, \nabla_{x} \Phi(x,t) \Big) 
= \frac{1}{2} \, \mean_{{p}_{\text{data}}(\bar{x})} \Big[ {\rho}(x \mid \bar{x},t) \, \Big( \gamma(x, \bar{x}, t) - \bar{\gamma}(x, \bar{x}, t) \Big) \Big]
\end{split}
\end{align}
is then equivalent to minimizing the KL divergence
$\KLD \big[ {\rho}_{\Phi}(x,t) \| \bar{\rho}(x,t) \big]$
between the flow-driven homotopy and the conditional homotopy.
\end{proposition}

\begin{proof}
Refer to Appendix \ref{Appendix:C}.
\end{proof}

Therefore, solving this density-weighted Poisson's equation corresponds to performing a homotopy matching ${{\rho}_{\Phi}} \equiv \bar{\rho}$. In the following section, we demonstrate that this homotopy matching gives rise to a Boltzmann energy expressed in terms of the potential energy $\Phi$ when the marginal homotopy $\bar{\rho}$ reaches its stationary equilibrium, thereby establishing a connection between our proposed potential flow framework and EBMs.

\subsection{Connections to Diffusion Process and Energy-Based Modeling} \label{sec: connections_diffusion_ebm}

In this section, we clarify the relationship between diffusion models and flow matching within the homotopy matching framework. Building on this insight, we establish a link between our proposed potential flow framework and energy-based modeling.

First, we present results from diffusion models. It has been outlined in \citet{NCSNpp} that the conditional density homotopy, represented by the Gaussian perturbation kernel ${\rho}(x \mid \bar{x},t) = \mathcal{N}\big(x; \mu(t) \, \bar{x}, \sigma(t)^2 I\big)$, characterizes a diffusion process governed by the following stochastic differential equation (SDE):
\begin{align} \label{eq:forward_time_SDE}
\begin{split} 
dx(t) = - f(t) \, x(t) \, dt + g(t) \, dW(t)
\end{split}
\end{align}
where $W(t) \in \mathbb{R}^n$ denote the standard Wiener process. Note that the time parameterization with respect to $t$ here is the reverse of the conventional parameterization used in diffusion models, where the diffusion process transitions from $x(1) \sim p_\text{data}(\bar{x})$ to $x(0) \sim q(x) = \mathcal{N}(0,\omega^2 I)$ as defined in Section \ref{ssect:LogHomotopy}.
In addition, the time-varying drift $f: [0, 1] \rightarrow \mathbb{R}$ and diffusion $g: [0, 1] \rightarrow \mathbb{R}$ coefficients are shown by \citet{EDM} to be given by
\begin{align} \label{eq:SDE_drift_diffusion}
\begin{split} 
&f(t) = - \frac{\dot{\mu}(t)}{\mu(t)}, \quad
g(t) = - \sqrt{\, 2 \, \sigma(t) \, \Big(\dot{\sigma}(t) + f(t) \, \sigma(t) \Big)}
\end{split}
\end{align}
where $\dot{\mu}(t)$ and $\dot{\sigma}(t)$ denote the time-derivatives.
It has also been shown in \citet{NCSNpp} that the following deterministic probability flow ODE:
\begin{align} \label{eq:forward_time_ODE}
\begin{split} 
\frac{dx(t)}{dt} &= - f(t) \, x(t) + \frac{1}{2} \, g(t)^2 \, \nabla_{x} \log \bar{\rho}(x,t)
\end{split}
\end{align}
results in the same marginal probability homotopy $\bar{\rho}(x,t)$ as the forward-time diffusion SDE (\ref{eq:SDE_drift_diffusion}).
Subsequently, we highlight the link between the diffusion process and the vector field modeled in flow matching.

\begin{proposition} \label{thm:proposition_4}
The conditional vector field in flow matching \citep{FlowMatching},  given by
\begin{align} \label{eq:conditional_vector_field}
\begin{split} 
\frac{dx(t)}{dt} = v(x \mid \bar{x},t) = \dot{\mu}(t) \, \bar{x} + \dot{\sigma}(t) \, \epsilon
\end{split}
\end{align}
with standard Gaussian noise $\epsilon \sim \mathcal{N}(0,I)$, satisfies the conditional probability flow ODE
governing the diffusion process conditioned on boundary condition $x(1) \sim p_\text{data}(\bar{x})$. 
It follows that the marginal vector field, given by the law of iterated expectation (tower property) $\mathbb{E}[U | X=x] = \mathbb{E}[\mathbb{E}[U \mid X=x,Y] \mid X=x]$:
\begin{align} \label{eq:marginal_vector_field}
\begin{split} 
\frac{dx(t)}{dt} = v(x,t) 
= \mathbb{E}_{p_{\text{data}}(\bar{x} \mid x)} \big[ v(x \mid \bar{x},t) \mid x \big] 
= \int_{\Omega} v(x \mid \bar{x},t) \, \frac{{\rho}(x \mid \bar{x},t) \, p_{\text{data}}(\bar{x})}{\bar{\rho}(x,t)} \, d{\bar{x}}
\end{split}
\end{align}
also satisfies the marginal probability flow ODE (\ref{eq:forward_time_ODE}).
\end{proposition}

\begin{proof}
Refer to Appendix \ref{Appendix:D}.
\end{proof}

Building on this result, we establish a connection between the proposed potential flow framework and EBMs. The following proposition demonstrates that homotopy matching, e.g., ${{\rho}_{\Phi}} \equiv \bar{\rho}$ leads to an energy-parameterized Boltzmann equilibrium.

\begin{proposition} \label{thm:proposition_5}
Given that the flow-driven homotopy ${\rho}_{\Phi}(x,t)$ matches the data-recovery marginal homotopy $\bar{\rho}(x,t)$, they exhibit the same Fokker–Planck dynamics. As the time-varying marginal density $\bar{\rho}(x,t)$ converges to its stationary equilibrium $\bar{\rho}_{\infty}(x)$, i.e., when $\frac{\partial \bar{\rho}(x,t)}{\partial t} \rightarrow 0$, the Fokker-Planck dynamics reach the Boltzmann distribution (\ref{eq:Boltzmann_distribution}), where the Boltzmann energy $\Phi_{B}(x)$ is defined as follows: 
\begin{align}
\begin{split} \label{eq:Boltzmann_energy}
{\Phi}_{B}(x) = \frac{4 \, {\Phi}_{\infty}(x) + f_{\infty} \, \|x\|^2}{g_{\infty}^2}
\end{split}
\end{align}
\end{proposition}
where ${\Phi}_{\infty}(x)$, $f_{\infty}$, and $g_{\infty}$ denote the steady-state potential energy, drift, and diffusion coefficients, respectively, associated with the stationary equilibrium.

\begin{proof}
Refer to Appendix \ref{Appendix:E}.
\end{proof}

On that note, we uncover the connection between the proposed VPFB framework and EBMs, demonstrating the validity of the potential energy as a parameterization of a Boltzmann energy. This holds provided that $\bar{\rho}$ converges to its stationary equilibrium and ${\rho}_{\Phi}$ learns to match these convergent dynamics. In the following section, we introduce a variational principle approach to solving the density-weighted Poisson equation (\ref{eq:unmarginalized_PDE_tractable}), thereby addressing the intractable homotopy matching problem.

\subsection{Variational Potential Energy Loss Formulation: Deep Ritz Method}\label{sec: variational_loss}

Solving the density-weighted Poisson's equation (\ref{eq:unmarginalized_PDE_tractable}) is particularly challenging in high-dimensional settings. Numerical approximation struggles to scale with higher dimensionality, as selecting suitable basis functions, such as in the Galerkin approximation, becomes increasingly complex \citep{Yang_2}. 
Similarly, a diffusion map-based algorithm demands an exponentially growing number of particles to ensure error convergence \citep{Diffusion_Map}. To address these challenges, we propose a variational loss function using the Deep Ritz method. This approach casts Poisson's equation as a variational problem compatible with stochastic gradient descent. Consequently, the proposed approach solves Eq. (\ref{eq:unmarginalized_PDE_tractable}), effectively aligning the flow-driven homotopy with the marginal homotopy. Directly solving Poisson's equation (\ref{eq:unmarginalized_PDE_tractable}) is challenging. Therefore, we first consider the following weak formulation:
\begin{align}
\begin{split} \label{eq:weak_formulation}
& \int_{\Omega} \, \frac{1}{2} \, \mean_{{p}_{\text{data}}(\bar{x})} \Big[ {\rho}(x \mid \bar{x},t) \, \big( \gamma(x, \bar{x}, t) - \bar{\gamma}(x, \bar{x}, t) \big) \Big] \, \Psi \, dx 
= \int_{\Omega} \nabla_{x} \cdot \Big( \bar{\rho}(x,t) \, \nabla_{x} \Phi(x,t) \Big) \bigg) \, \Psi \, dx
\end{split}
\end{align}
This PDE must hold for all differentiable trial functions $\Psi$. In the following proposition, we introduce a variational loss function that is equivalent to solving this weak formulation of the density-weighted Poisson's equation.

\begin{proposition} \label{thm:proposition_6}
The variational problem of minimizing the following loss functional:
\begin{align} \label{eq:variational_functional}
\begin{split}
\mathcal{L}(\Phi,t) = &\; \cov_{{\rho}(x \mid \bar{x},t) \, {p}_{\text{data}}(\bar{x})} \Big[ \Phi(x,t) , \gamma(x, \bar{x}, t) \Big] 
+ \mean_{\bar{\rho}(x,t)} \Big[ \big\| \nabla_{x} \Phi(x,t) \big\|^{2} \Big]
\end{split}
\end{align}
with respect to the potential energy $\Phi$, 
is equivalent to solving the weak formulation (\ref{eq:weak_formulation}) of the density-weighted Poisson's equation (\ref{eq:unmarginalized_PDE_tractable}).
Here, $\|\cdot\|$ denotes the Euclidean norm, and $\cov$ denotes the covariance.
Furthermore, the variational problem (\ref{eq:variational_functional}) admits a unique solution $\Phi \in \mathcal{H}^1_0(\Omega; \rho)$ if the marginal homotopy $p$ satisfy the Poincar\'e inequality:
\begin{align} \label{eq:Poincare_inequality}
\begin{split}
\mean_{\bar{\rho}(x,t)} \Big[ \big\| \nabla_{x} \Phi(x,t) \big\|^{2} \Big] \geq \eta \, \mean_{\bar{\rho}(x,t)} \Big[ \big\| \Phi(x,t) \big\|^{2} \Big]
\end{split}
\end{align}
for some positive scalar constant $\eta > 0$ (spectral gap).
\end{proposition}

\begin{proof}
Refer to Appendix \ref{Appendix:F}.
\end{proof}

\begin{remark}
The integration by parts in (\ref{eq:divergence_theorem}) and (\ref{eq:integration_by_parts}) require that the marginal density $\bar{\rho}(x)$ vanishes on the boundary $\partial \Omega$ of some open, bounded domain $\Omega \subset \mathbb{R}^n$, so that the boundary integral $\int_{\partial \Omega} \bar{\rho}(x) \, (\nabla_{x} \Phi \cdot \hat{n}) \, dx = 0$ holds. In standard implementations, although the training data $\bar{x}$ are typically normalized to lie within $[-1,1]^n$, we may define the perturbed samples as $x \in \Omega \subset \mathbb{R}^n$, where $\Omega$ is chosen to contain the support of the data distribution. Accordingly, the open bounded domain $\Omega$ can be defined sufficiently large so that the conditional homotopy ${\rho}(x \mid \bar{x},t)$ approaches zero at the boundary $\partial \Omega$. Since ${\rho}(x \mid \bar{x},t)$ is a Gaussian perturbation kernel, it decays exponentially and is effectively negligible near the boundary, thereby satisfying the required condition. As a result, the marginal distribution $\bar{\rho}(x,t)$ also vanishes at $\partial \Omega$, ensuring the validity of the integration by parts required to formulate both Proposition \ref{thm:proposition_3} Proposition \ref{thm:proposition_6}.
\end{remark}

Overall, Propositions \ref{thm:proposition_3} and \ref{thm:proposition_6} recast the intractable problem of minimizing the KL divergence between the flow-driven homotopy and the marginal homotopy as an equivalent variational problem of solving the loss function (\ref{eq:variational_functional}). By optimizing the potential energy with respect to this loss and transporting the prior samples through the ODE (\ref{eq:particle_flow_ODE}), the prior particles evolve along a trajectory that aligns with the marginal homotopy.
In particular, the covariance loss here plays an important role 
by ensuring that the normalized innovation (residual sum of squares) is inversely proportional to the potential energy. As a result, the energy-generated velocity field $\nabla_{x} \Phi$ consistently points in the direction of greatest potential ascent, thereby driving the flow of prior particles towards high likelihood regions. Given that homotopy matching is performed over the entire time horizon, we apply stochastic integration to the loss function over time, where $t \sim \mathcal{U}(0,t_\text{end})$ is drawn from the uniform distribution.


\subsection{Training Implementation}\label{implementation}

In our implementation, we adopt the Optimal Transport Flow Matching (OT-FM) framework \citep{FlowMatching} for training, where it corresponds to the SDE parameterization $f(t) = - \frac{1}{t}$, $g(t) = \sqrt{\frac{2 \, (1-t)}{t}}$ as outline in \citep{DiffusionELBO}, or equivalently $\alpha(t) = \frac{\omega^2}{1 - t}$, $\beta(t) = \frac{\nu^2 \, t}{(1 - t)^2}$ for the log-homotopy transformation derived in (\ref{eq:log-homotopy_f}).
To establish a Boltzmann equilibrium, we further require $\frac{\partial \bar{\rho}(x,t)}{\partial t} \rightarrow 0$ so that the time-varying marginal density $\bar{\rho}(x,t)$ converges to the stationary Boltzmann distribution. However, the Gaussian perturbation kernel $\rho(x \mid \bar{x}, t) = \mathcal{N}\big(\mu(t) \, \bar{x}, \sigma(t)^2 I\big)$ employed by the flow-based probability paths is defined only over a finite time interval $t \in [0,t_{\text{max}}]$. Additionally, the marginal homotopy $\bar{\rho}(x,t)$ is not guaranteed to reach equilibrium within this prescribed time window. 

To resolve these limitations of the flow-based probability paths, we explicitly enforce stationarity in our training implementation, by imposing a steady-state equilibrium $p_{\infty}(x) = \bar{\rho}(x, t \geq t_{\text{max}})$ beyond some cutoff time $t_{\text{max}} < t_{\text{end}}$ close to the terminal time. This steady-state equilibrium $p_{\infty}(x) \equiv p_B(x)$ thus corresponds to the stationary Boltzmann distribution, parameterized by the energy function derived in (\ref{eq:Boltzmann_energy}) with $f_{\infty} = f(t_{\text{max}})$ and $g_{\infty} = g(t_{\text{max}})$. Given that a steady-state equilibrium is enforced via $p_{\infty}(x) = \bar{\rho}(x, t \geq t_{\text{max}}) \approx p_{\text{data}}(\bar{x})$, the stationary Boltzmann distribution approximates the true data likelihood by design. 

Finally, our VPFB loss function is implemented as follows:
\begin{align} \label{eq:VPFB_loss_integral}
\begin{split}
\mathcal{L}^{\mathrm{VPFB}}(\Phi) &= \int_{0}^{t_\text{end}} \mathcal{L}(\Phi,t) \, dt = \mean_{\mathcal{U}(0,t_\text{end})} \big[ \mathcal{L}(\Phi,t) \big]
\end{split}
\end{align}
where
\begin{align} \label{eq:VPFB_loss}
\begin{split}
\mathcal{L}(\Phi,t) = &\; \cov_{{\rho}(x \mid \bar{x},t) \, {p}_{\text{data}}(\bar{x})} \Big[ \Phi(x,t) , w(t) \,  \gamma(x, \bar{x}, t) \Big]
\;- \frac{\nabla_x \Phi(x,t) \cdot v(x \mid \bar{x}, t)}{\big\|\nabla_x \Phi(x,t)\big\| \, \big\|v(x \mid \bar{x}, t)\big\|} \\
&\;+ \mean_{{\rho}(x \mid \bar{x},t) \, {p}_{\text{data}}(\bar{x})} \Big[ \big\| \nabla_{\!(x,t)} \Phi(x,t)  \big\|^{2} + \eta \, \big\| \Phi(x,t)  \big\|^{2} \Big]
\end{split}
\end{align}
Here, we incorporate an additional cosine distance between the potential gradient $\nabla_{x} \Phi$ and the conditional vector field in (\ref{eq:conditional_vector_field}) to the loss function. While this cosine distance does not influence the learning of the potential energy’s magnitude (magnitude learning is entirely supervised by the covariance loss), it enforces directional alignment between the gradient and the vector field. To further enforce convergence toward a steady-state potential $\Phi_{\infty}(x)$, we additionally encourage quasi-static dynamics by minimizing the Euclidean norm of the time derivative $\left|\frac{\partial \Phi}{\partial t}\right|^2$ alongside the gradient norm during training. Also, a weighting $w(t) = (1 - t)^{\kappa}$ with decay exponent $\kappa > 1$ is applied to the innovation term to balance the covariance loss across time to stabilize training.

Considering that the marginal homotopy may not satisfy the Poincaré inequality (\ref{eq:Poincare_inequality}), we include the right-hand side of this inequality in the loss function to enforce the uniqueness of the minimizer. To empirically validate the existence of a positive Poincaré constant $\eta$, Figure \ref{fig:poincare} plots the ratio between the mean gradient norm ${\mathbb{E}} \big[\| \nabla_x \Phi \|^{2} \big]$ and the mean energy norm ${\mathbb{E}} \big[\| \Phi \|^{2} \big]$ over training iterations on CIFAR-10, without applying the additional Poincaré regularization loss. It shows that the ratio is bounded below by $\eta = 6.81 \times 10^{-5}$, thereby confirming the existence of a positive Poincaré constant during training. 

Nonetheless, our experiments indicate that the existence and magnitude of such an unenforced Poincaré constant vary across different neural architectures. 
For completeness, we incorporate the Poincaré regularization with a small $\eta$ for both the WideResNet and U-Net models, which we fine-tune during training for optimal results. The cutoff time $t_\text{max}$, terminal time $t_\text{end}$, decay exponent $\kappa$, and spectral gap constant $\eta$ are hyperparameters to be determined during training.
Algorithm~\ref{algo:VPFB_training} summarizes the training procedure of our proposed VPFB framework.


\begin{figure}[t]
\centering
\includegraphics[width=1.0\columnwidth]{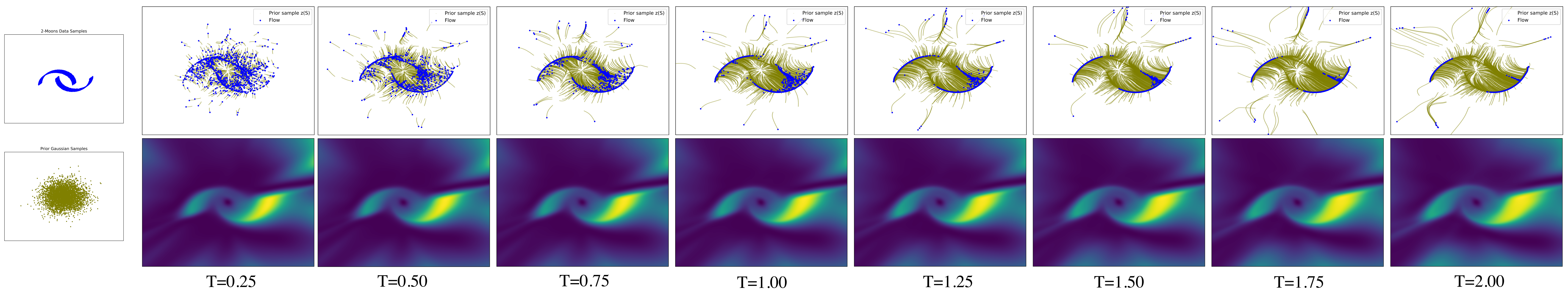}
\caption{2D potential flow. Top: Sample trajectories from the Gaussian prior noise distribution (black) to the target 2-Moons distribution (blue), driven by the potential energy $\Phi(x,t)$ and sampled using an ODE solver. Bottom: Time evolution of the learned potential energy landscape $\Phi(x,t)$.}
\label{fig:density_potential}
\end{figure}

\begin{figure}[t]
\centering
\includegraphics[width=1.0\columnwidth]{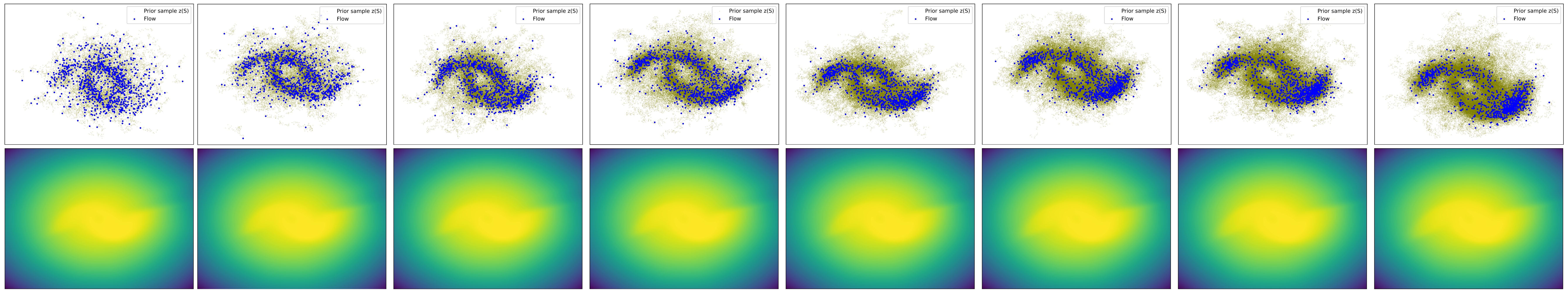}
\caption{2D Boltzmann density estimation. Top: Sample trajectories from the Gaussian prior noise distribution (black) to the target 2-Moons distribution (blue), driven by the Boltzmann energy and sampled via SGLD. Middle: Visualization of the log-density estimation (up to an additive constant) $\log p_B (x) = {\Phi}_{B}(x)$ parameterized by Boltzmann energy.}
\label{fig:density_Boltzmann}
\end{figure}

\section{Experiments}

In this section, we validate the energy-based generative modeling capabilities of VPFB across several key tasks. Section \ref{sec:2Ddensityestimation} explores 2D density estimation. Section \ref{sec:unconditional_image} presents the unconditional generation and spherical interpolation results on CIFAR-10 and CelebA. 
Section \ref{sec:mode_evaluation} evaluates mode coverage and model generalization through energy histograms of train and test data and the nearest neighbors of generated samples. 
Section \ref{sec:ood_detection} examines unsupervised OOD detection performance on various datasets. 
Section \ref{sec:long_run} verifies the convergence of long-run ODE samples to a Boltzmann equilibrium. 
Additional results on ablation study and computational efficiency are provided in Appendix \ref{sec:additional_results}.
Additional discussions of the results are also provided in Appendix \ref{sec:additional_discussions}.
Finally, implementation details, including architecture, training, numerical solvers, datasets, and FID evaluation, are provided in Appendix \ref{sec:experimental_details}.

\subsection{Density Estimation on 2D Data}\label{sec:2Ddensityestimation}
To verify the convergence properties of the potential energy and to assess the validity of the Boltzmann energy (\ref{eq:Boltzmann_energy}), we conduct density estimation on 2D synthetic datasets. Specifically, we learn a potential flow that transforms an unimodal Gaussian prior distribution into a 2-Moons target distribution. 
Figure \ref{fig:density_potential} shows the sample trajectories driven by the potential flow $dx(t) = \nabla_{x} \Phi(x, t) \, dt$, obtained via the deterministic Euler solver. Figure \ref{fig:density_Boltzmann} presents the sample trajectories and density estimation of the Boltzmann distribution $p_B \propto e^{\Phi_{B}(x)}$, obtained via the Stochastic Gradient Langevin Dynamics (SGLD). Notably, both the potential energy $\Phi(x)$ and the Boltzmann energy $\Phi_{B}(x)$ exhibit stable convergence toward their steady-state equilibrium. Furthermore, the results indicate that the estimated Boltzmann density closely aligns with the ground-truth 2-Moons distribution.
These results highlight the effectiveness of our variational principle approach in learning the Boltzmann stationary distribution through homotopy matching against the stationary-enforced marginal $p_{\infty}(x)$.

Nonetheless, a standard formulation of the forward-time SDE (\ref{eq:forward_time_SDE}), or equivalently, the marginal probability flow ODE (\ref{eq:forward_time_ODE}), is valid only in the case of a unimodal Gaussian prior, e.g., $q(x) = {\rho}(x \mid \bar{x}, t=0) = \mathcal{N}(0, \omega^2 I)$, as discussed in \cite{DiffusionELBO}. This assumption underpins the consistency of the Fokker–Planck dynamics with the continuous-time diffusion framework, ensuring the validity of the stationary Boltzmann energy in the limit. We acknowledge this limitation of our current framework.
As a direction for future work, we propose extending the forward-time SDE or ODE formulation of continuous-time diffusion to be admissible for more general prior distributions, such as mixtures of Gaussians or learned priors, to accommodate multi-modal data while maintaining consistency with our proposed energy-based framework.

\begin{figure}[t]
\centering
\includegraphics[width=1.0\columnwidth]{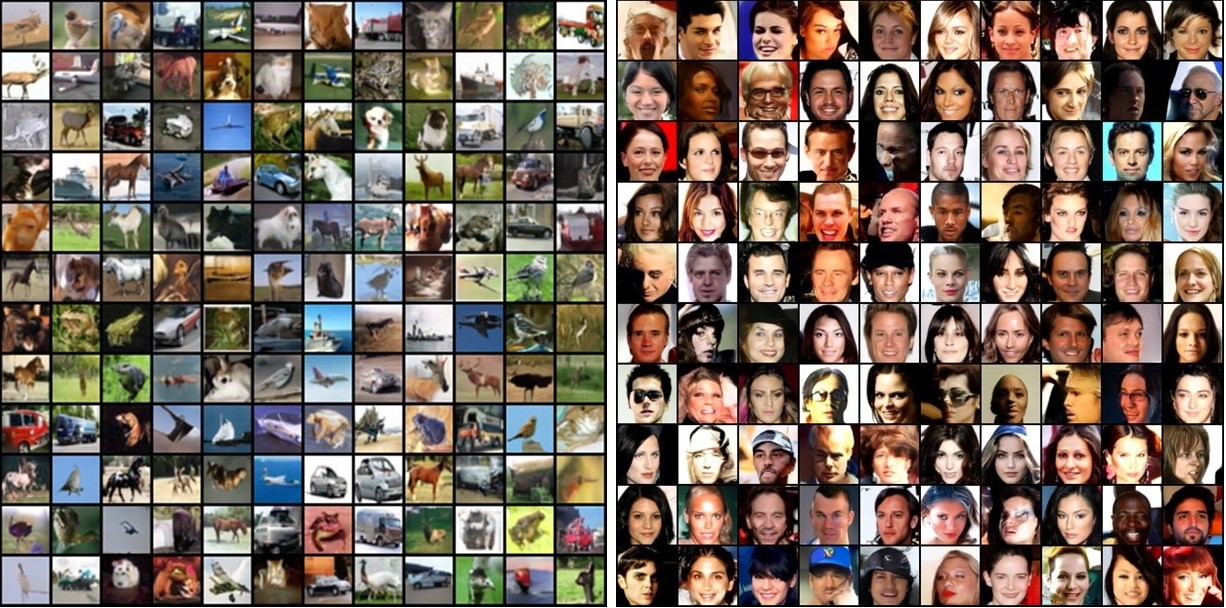}
\caption{Uncurated and unconditional samples generated for CIFAR-10 (left) and CelebA (right).}
\label{fig:cifar10_small}
\end{figure}

\subsection{Unconditional Image Generation}\label{sec:unconditional_image}

For image generation, we consider three VPFB model variants: an autonomous (independent of time) energy model $\Phi(x)$ parameterized by  \citet{wideresnet}, and a time-varying energy model $\Phi(x, t)$ parameterized by U-Net \citep{Unet}.
Figure \ref{fig:cifar10_small} shows the uncurated and unconditional image samples generated using the time-varying energy model on CIFAR-10 $32\times 32$ and CelebA $64 \times 64$. 
The generated samples are of decent quality and resemble the original datasets, despite not having the highest fidelity as achieved by state-of-the-art models. Table \ref{tab:fid_cifar10} summarizes the quantitative evaluations of our framework in terms of FID \citep{FID} scores on the CIFAR-10. In particular, the VPFB models achieved FID scores competitive to existing generative models. 
Figures \ref{fig:cifar10_big} and \ref{fig:celeba_big} show additional uncurated samples of unconditional image generation on CIFAR-10 and CelebA, respectively.

\subsection{Image Interpolation and Compositional Generation}\label{sec:compositional_image}
\begin{figure}[t]
\centering
\includegraphics[width=1.0\columnwidth]{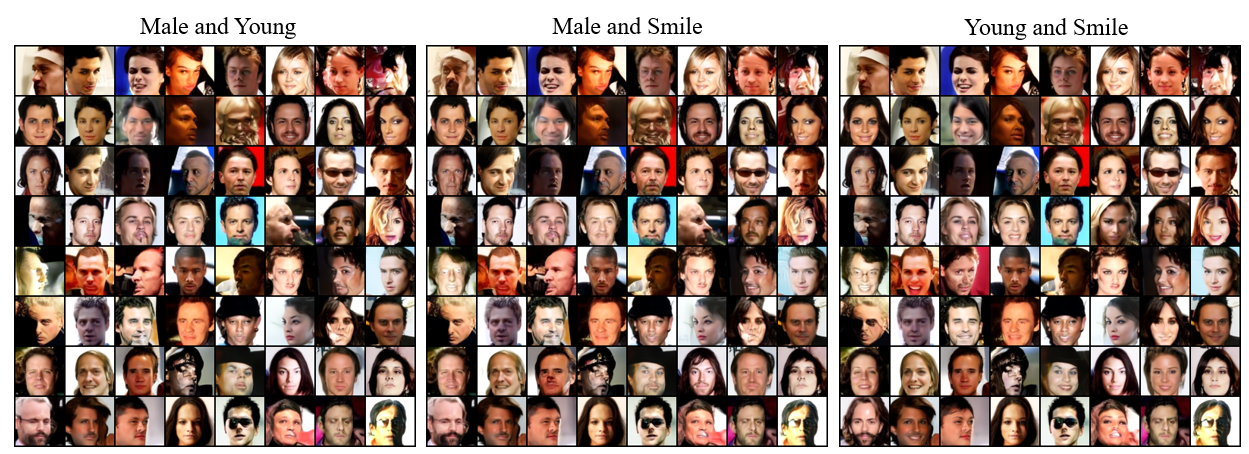}
\caption{Compositional and conditional CelebA samples generated based on three attribute pairs.}
\label{fig:compositionality}
\end{figure}

To achieve smooth and semantically coherent image interpolation, we perform spherical interpolation between two Gaussian noises and subsequently apply ODE sampling to the interpolated noises.
Figures \ref{fig:cifar10_interp_big} and \ref{fig:celeba_interp_big} show additional interpolation results on CIFAR-10 and CelebA, respectively. 
For compositional sample generation, we first train a class-conditioned energy model $\Phi(x, c)$, and then sample by averaging the conditional energies across selected classes.
Figure \ref{fig:compositionality} presents compositional generation results conditioned on composite CelebA attributes, specifically (\textit{Male, Young}), (\textit{Male, Smile}), and (\textit{Young, Smile}).
However, certain compositional samples show limited variation across attribute pairs, suggesting that incorporating composition weights could improve attribute-specific conditioning.

\begin{figure}[t]
\centering
\includegraphics[width=0.75\columnwidth]{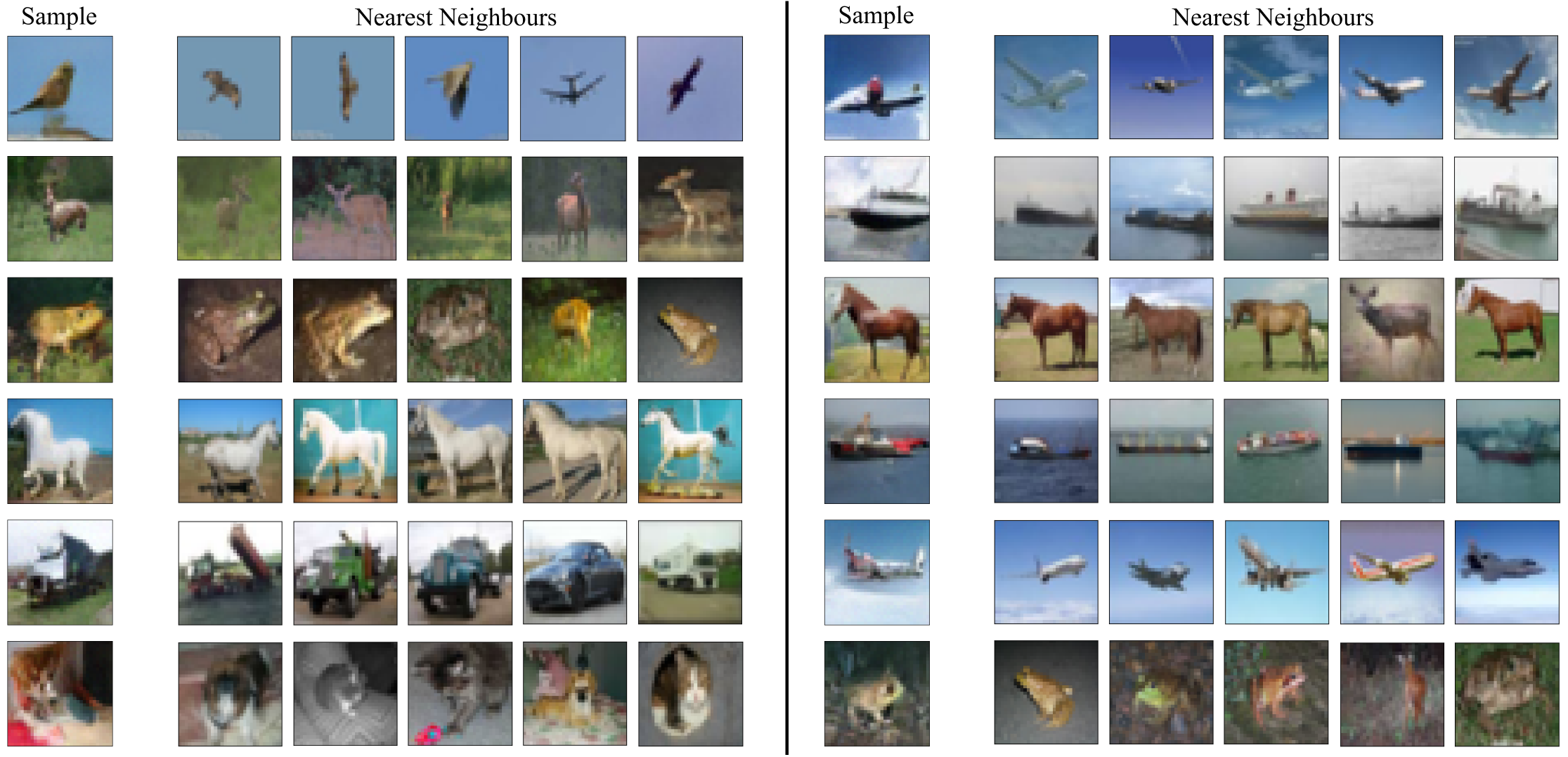}
\caption{Generated CIFAR-10 samples and their five nearest neighbors in train set based on pixel distance.}
\label{fig:knn_cifar10}
\end{figure}

\subsection{Model Generalization and Mode Evaluation} \label{sec:mode_evaluation}
To evaluate the generalization capability of the VPFB model, Figure \ref{fig:knn_cifar10} presents the nearest neighbors of the generated samples in the CIFAR-10 training set. The results show that nearest neighbors differ significantly from the generated samples, suggesting that our model does not overfit the training data and generalizes well to the underlying data distribution. 
To validate the mode coverage and over-fitting ability, Figure \ref{fig:histogram_cifar10} presents a histogram of the energy outputs for both the CIFAR-10 training and test datasets. The histogram shows that the learned energy model assigns similar energy values to images from both sets. This indicates that the VPFB model generalizes well to unseen test data while maintaining broad mode coverage of the training distribution.

\begin{table*}[t]
\small
\caption{FID scores on unconditional CIFAR-10 image generation.}
\vspace{1em}
\label{tab:fid_cifar10}
\centering
\setlength{\aboverulesep}{0pt}
\setlength{\belowrulesep}{1pt}
\begin{tabular}{p{0.3825\textwidth} p{0.06\textwidth} | p{0.3825\textwidth} p{0.06\textwidth}}
\toprule
\textbf{Energy-based Models} & {\textbf{FID $\downarrow$}} & \textbf{Other Likelihood-based Models} & {\textbf{FID $\downarrow$}} \\
\midrule
EBM-IG \citep{IGEBM} & 38.2 & ResidualFlow \citep{residualflow} & 47.4  \\
EBM-FCE \citep{Flow_Contrastive} & 37.3 & Glow \citep{glow} & 46.0 \\
CoopVAEBM \citep{CoopVAEBM} & 36.2 & DC-VAE \citep{dc-vae} & 17.9  \\
\cmidrule(r){3-4}
CoopNets \citep{CoopNets} & 33.6 & \textbf{GAN-based Models} & \\
\cmidrule(r){3-4}
Divergence Triangle \citep{divergence_triangle} & 30.1 & SN-GAN \citep{Miyato} & 21.7 \\
VERA \citep{VERA} & 27.5 & SNGAN-DDLS \cite{DDLS} & 15.4 \\
EBM-CD \citep{IGEBMpp}  & 25.1 & BigGAN \citep{BigGAN} & 14.8  \\
\cmidrule(r){3-4}
GEBM \citep{GEBM}  & 19.3 & {\textbf{Score-based and Diffusion Models}} & \\
\cmidrule(r){3-4}
HAT-EBM \citep{HatEBM} & 19.3 & NCSN-v2 \citep{NCSNv2} & 10.9 \\
CF-EBM \citep{CFEBM} & 16.7 & DDPM Distil \citep{DDPM_distil} & 9.36 \\
CoopFlow \citep{CoopFlow} & 15.8 & DDPM \citep{DDPM} & 3.17 \\
VAEBM \citep{VAEBM} & 12.2 & NCSN++ \citep{NCSNpp} & 2.20 \\
\cmidrule(r){3-4}
DRL \citep{Diffusion_Recovery} & 9.58 & {\textbf{Flow-based Models}} & \\
\cmidrule(r){3-4}
CLEL \citep{CLEL} & 8.61 & Action Matching \citep{ActionMatching} & 10.0 \\
DDAEBM \citep{DDAEBM} & 4.82 & Flow Matching \citep{FlowMatching} & 6.35 \\
CDRL \citep{Cooperative_Diffusion_Recovery} & 3.68 & Rectified Flow \citep{RectifiedFlow} & 4.85 \\
\cmidrule(l){1-2}
VPFB (Autonomous) & \textbf{14.5} & DSBM \citep{BridgeMatching} & 4.51 \\
VPFB (Time-varying) & \textbf{6.72} & PFGM \citep{PFGM} & 2.35 \\
\bottomrule
\end{tabular}
\end{table*}

\begin{table}[t]
\small
\caption{AUROC scores $\uparrow$ for OOD detection on several datasets.}
\vspace{1em}
\label{tab:auroc}
\centering
\setlength{\aboverulesep}{0pt}
\setlength{\belowrulesep}{1pt}
\resizebox{\columnwidth}{!}{
\begin{tabular}{lllll} 
\toprule
\textbf{Models} & 
\textbf{\begin{tabular}[c]{@{}c@{}}CIFAR-10 Interpolation\end{tabular}} & 
\textbf{CIFAR-100} & 
\textbf{CelebA} & 
\textbf{SVHN}\\ 
\midrule
PixelCNN \citep{pixelcnn} & 0.71 & 0.63 & - & 0.32 \\
GLOW \citep{glow} & 0.51 & 0.55 & 0.57 & 0.24 \\
NVAE \citep{NVAE} & 0.64 & 0.56 & 0.68 & 0.42 \\
EBM-IG \citep{IGEBM} & 0.70 & 0.50 & 0.70 & 0.63 \\
VAEBM \citep{VAEBM} & 0.70 & 0.62 & 0.77 & 0.83 \\
CLEL \citep{CLEL} & 0.72 & 0.72 & 0.77 & 0.98 \\
DRL \citep{Diffusion_Recovery} & - & 0.44 & 0.64 & 0.88 \\ 
CDRL \citep{Cooperative_Diffusion_Recovery} & 0.75 & 0.78 & 0.84 & 0.82 \\ 
\midrule
VPFB (Ours) & 0.78 & 0.67 & 0.84 & 0.61 \\ 
\bottomrule
\end{tabular}
} 
\end{table}

\subsection{Out-of-Distribution Detection} \label{sec:ood_detection}
Given that the potential flow corresponds to a stationary Boltzmann distribution, the Boltzmann energy ${\Phi}_{B}$ from (\ref{eq:Boltzmann_energy}) can be used to distinguish between in-distribution and OOD samples based on their assigned energy values.
Specifically, the potential energy model trained on the CIFAR-10 training set assigns energy values to both in-distribution samples (CIFAR-10 test set) and OOD samples from various other image datasets. We evaluate OOD detection performance using the AUROC metric, where a higher score reflects better model’s efficacy in accurately assigning lower energy values to OOD samples.
Table \ref{tab:auroc} compares the AUROC scores of VPFB with those of various likelihood-based and EBMs. The results show that our model performs exceptionally well on interpolated CIFAR-10 and CelebA $32 \times 32$ while achieving moderate performance on CIFAR-100 and SVHN.

\subsection{Long-Run Steady-State Equilibrium} \label{sec:long_run}

Figure \ref{fig:longrun_ODE_sample_cifar10_celeba} illustrates long-run ODE sampling over an extended time horizon $t \in [0, 20]$ using the autonomous energy model parameterized by WideResNet. Additionally, Figure \ref{fig:longrun_ODE_sample_celeba} illustrates long-run ODE sampling using the time-varying energy model parameterized by U-Net. The results indicate a similar deterioration in image quality over extended time periods, albeit to a greater extent compared to the autonomous model. Figure \ref{fig:longrun_ODE_plot_celeba} plots the mean gradient norm $\mathbb{E} \big[ \|\nabla_{x} \Phi\|^{2} \big]$ and the mean energy norm $\mathbb{E} \big[ \| \Phi \|^{2} \big]$, neither of which exhibit convergence.
These results are consistent with those observed in EBMs trained using non-convergent short-run MCMC \citep{NonConvergentMCMC,Nijkamp_2020}. This issue arises from the inherent difficulty neural network models face in learning complex energy landscapes in high-dimensional spaces. Regions that remain unseen during training can correspond to poorly modeled areas of the energy landscape, often resulting in the emergence of sharp local minima. Consequently, ODE-based sampling may become trapped in these local minima, leading to mode collapse and poor mixing, which manifest as visual artifacts such as excessive saturation and loss of background details. 

To resolve these issues, we replace the deterministic ODE solver with the conventional SGLD sampler for image generation, enabling sampling from the Boltzmann energy via
$x_{t+1} = x_t + \Delta_t \, \nabla_x \Phi_{B}(x_t) + \sqrt{2 \, \Delta_t} \, \epsilon$
where $\epsilon \sim \mathcal{N}(0, \lambda^2 I)$ denotes isotropic Gaussian noise with temperature scale $\lambda$ (standard deviation), and $\Delta_t$ is the step size. The injected stochasticity from the diffusive noise in SGLD facilitates escape from local minima and enhances mixing efficiency during sampling. As shown in Figures \ref{fig:longrun_SDE_sample_celeba} and \ref{fig:longrun_SDE_sample_cifar10}, SGLD mitigates mode collapse and the long-run image samples converge well to the stationary equilibrium. Furthermore, Figures \ref{fig:longrun_SDE_plot_celeba} and \ref{fig:longrun_SDE_plot_cifar10} demonstrate that the gradient norm converges to zero, while the energy norm asymptotically stabilizes, indicating steady-state thermalization. 
These SDE-based sampling results confirm that equilibrium convergence is achievable with a stochastic sampler. Nonetheless, the temperature scale $\lambda$ must be carefully tuned to balance convergence speed and sample quality. Moreover, our experiments show that ODE-based sampling consistently yields better FID scores, potentially due to the deterministic nature of the proposed potential flow and the straightness of the linearly interpolated OT-FM trajectories, which contribute to sharper and more consistent sample generation.

\section{Conclusion}
We propose VPFB, a novel energy-based potential flow framework designed to reduce the computational cost and instability typically associated with EBM training. Empirical results demonstrate that VPFB outperforms several existing EBMs in unconditional image generation and achieves competitive performance in OOD detection, highlighting its versatility across diverse generative modeling tasks. Despite these promising results, future work will aim to refine the training strategy to improve scalability to higher-resolution images and other data modalities, while addressing the limitations outlined in this work. Additionally, exploring generative models that inherently incorporate Neumann boundary conditions into the design of their blurring perturbation kernels \citep{HeatDissipation,BlurringDiffusion,SoftDiffusion} presents a promising direction for improving energy landscape modeling and enhancing sample diversity without incurring the computational burden of long-run MCMC sampling.

\subsubsection*{Broader Impact Statement}
Generative models represent a rapidly growing field of study with overarching implications in science and society. Our work proposes a new generative model designed for efficient data generation and OOD detection, with potential applications in fields such as medical imaging, entertainment, and content creation. However, as with any powerful technology, generative models come with substantial risks, including the potential misuse in creating deepfakes or misleading content that could undermine social security and trust. Given this dual-use nature, it is essential to implement safeguards, such as classifier-based guidance, to prevent the generation of biased or harmful content. 
Moreover, generative models are vulnerable to backdoor adversarial attacks and can inadvertently amplify biases present in the training data, reinforcing social inequalities. Although our work uses standard datasets, it is important to address how such biases are handled. We are actively exploring methods to identify and mitigate biases during both the training and generation phases. This includes employing fairness-aware training algorithms and evaluating the model’s output for biased patterns. One potential solution is incorporating privacy-preserving encryption techniques to safeguard sensitive data and ensure that generative models do not expose private information. Furthermore, while this work demonstrates the potential benefits of generative models, the ethical concerns surrounding their deployment must be considered. Addressing these issues will require ongoing collaboration to develop frameworks for responsible use, including transparency, model interpretability, and robust safeguards against malicious applications. By proactively engaging with these ethical concerns, the broader community can contribute to the responsible advancement of generative modeling technologies.




\bibliography{ref}
\bibliographystyle{tmlr}

\newpage
\appendix

\section{Additional Results}
\label{sec:additional_results}

In this section, we present additional experiments that further validate the effectiveness and efficiency of the proposed VPFB framework. We first conduct an ablation study to evaluate the contribution of key loss components and architectural choices, demonstrating their impact on both FID performance and convergence to the Boltzmann equilibrium. Then, we compare the computational efficiency of VPFB against recent EBM baselines, highlighting its advantages in training and inference time while maintaining competitive generative performance across model variants.

\subsection{Ablation Study}

To isolate and quantify the impact of individual components in the proposed VPFB loss function. Table \ref{tab:loss_configurations} presents an ablation study conducted on a smaller (VPFB-Base) model and a reduced training batch size to accelerate training. Notably, the FID scores increase without (B) the covariance loss and (C) the cosine distance gradient alignment, indicating that these loss components are essential to the VPFB training. We note that since (B) learns only the normalized gradient and not the energy magnitude, it requires careful tuning of denormalization during ODE sampling.

\begin{figure}[t]
\centering
\includegraphics[width=0.75\columnwidth]{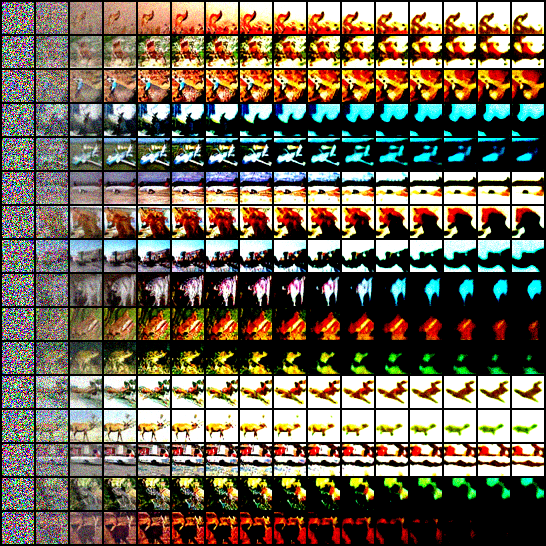}
\caption{Long-run SGLD sampling using the Boltzmann energy on CIFAR-10 for loss configuration (D).}
\label{fig:longrun_SDE_sample_cifar10_D}
\end{figure}

\begin{figure}[t]
\centering
\includegraphics[width=1.0\columnwidth]{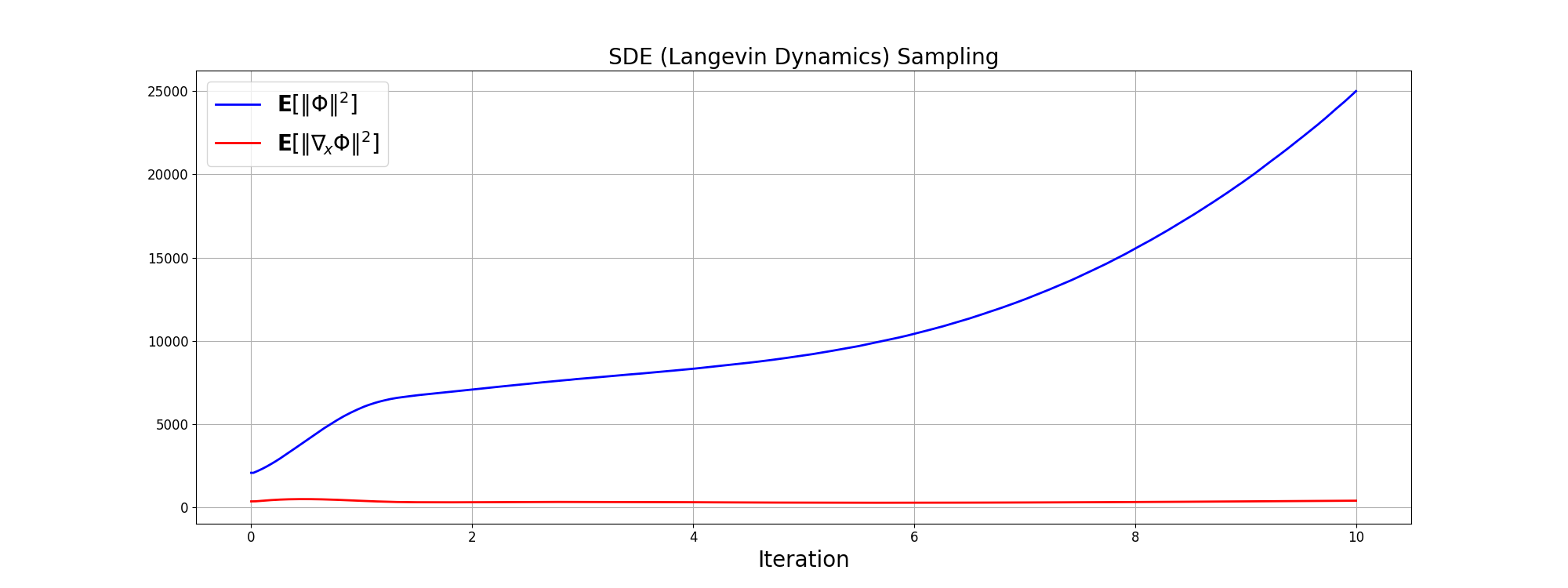}
\caption{Validation of the convergence of gradient norm and energy norm in long-run SGLD sampling with loss configuration (D) on CIFAR-10.}
\label{fig:longrun_SDE_plot_cifar10_D}
\end{figure}

\begin{figure}[t]
\centering
\includegraphics[width=0.75\columnwidth]{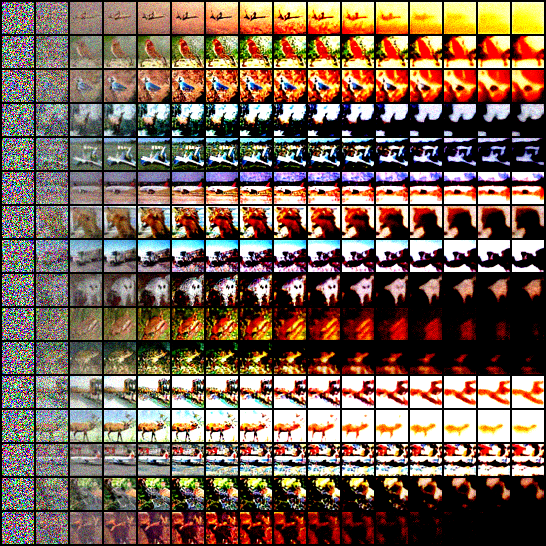}
\caption{Long-run SGLD sampling using the Boltzmann energy with loss configuration (E) on CIFAR-10.}
\label{fig:longrun_SDE_sample_cifar10_E}
\end{figure}

\begin{figure}[t]

\centering
\includegraphics[width=1.0\columnwidth]{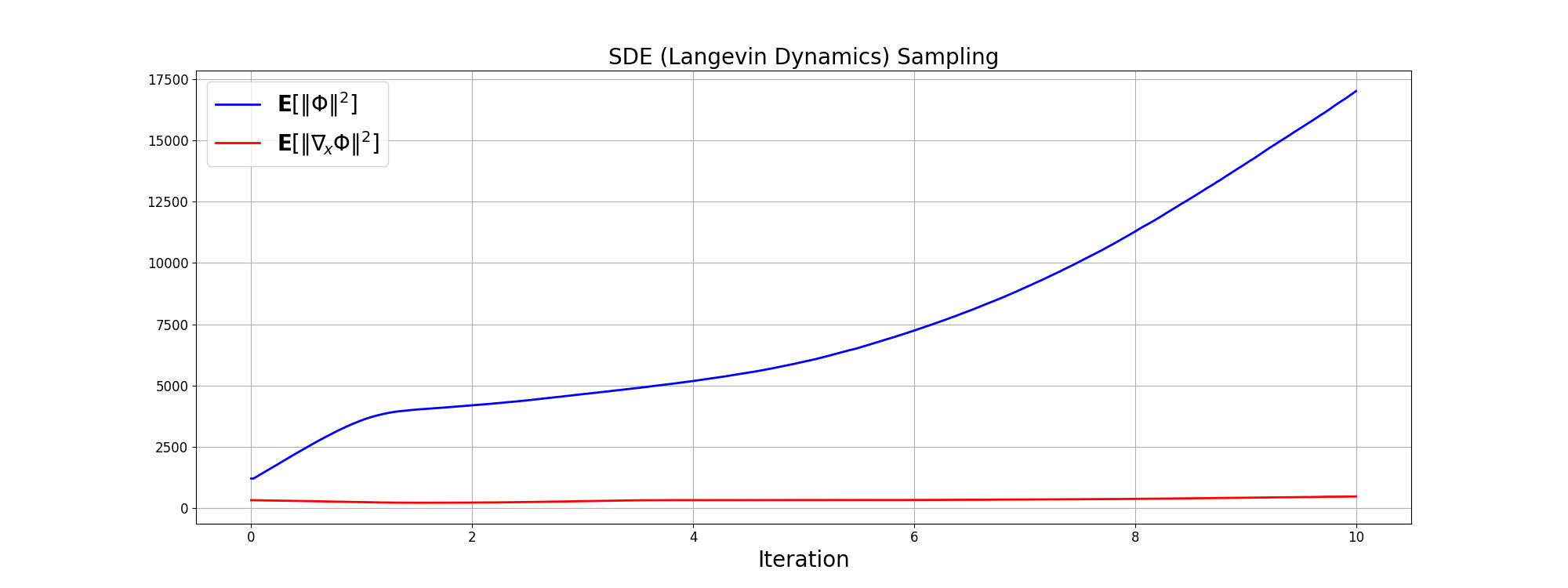}
\caption{Validation of the convergence of gradient norm and energy norm in long-run SGLD sampling with loss configuration (E) on CIFAR-10.}
\label{fig:longrun_SDE_plot_cifar10_E}
\end{figure}

Subsequently, (D) replaces the cosine distance with inner product, and (E) replaces the entire VPFB loss with the flow matching loss of \citet{FlowMatching}. Although these loss configurations yield better FID performances, the sampling results and norm plots in Figures \ref{fig:longrun_SDE_sample_cifar10_D} - \ref{fig:longrun_SDE_plot_cifar10_E} show that neither of these configurations achieves Boltzmann equilibrium under SGLD sampling. Removing the cosine distance in (D) eliminates the scale invariance of cosine similarity, leading to large variations in gradient magnitudes that disrupt the stability of energy required for steady-state convergence. Similarly, the flow matching loss in (E) does not inherently enforce Boltzmann stationarity. Applying the same weighting $w(t)$ to the flow matching loss would halt the learning of the gradient field cutoff time $t_{\text{max}}$. 

For these reasons, we do not adopt the loss configurations (D) and (E) in our loss framework, despite the lower FID scores. In contrast, the variational nature of the covariance loss allows it to enforce that the energy values $\Phi(x,t)$ remain stationary near $t=1$ without impeding learning. This covariance loss is fundamental as it corresponds to the Fokker-Planck dynamics underlying a probability path. More importantly, its adaptability to weighting ensures a proper establishment of the stationary Boltzmann distribution. Finally, by (F) incorporating a larger VPFB-Base model and (G) increasing the training batch size, we obtain improved FID scores with (A) the proposed VPFB loss function.


\begin{table}[b]
\small
\caption{Ablation study across different training losses and model configurations.}
\label{tab:loss_configurations}
\centering
\setlength{\aboverulesep}{0pt}
\setlength{\belowrulesep}{1pt}
\resizebox{\columnwidth}{!}{%
\begin{tabular}{lllll} 
\toprule
\textbf{Loss Configuration} & \textbf{Model Variant} & \textbf{Parameter Count} & \textbf{Batch Size} & \textbf{FID $\downarrow$} \\
\midrule
(A) VPFB loss & Base & 38.3M & 64 & 9.45 \\
(B) VPFB loss without covariance & Base & 38.3M & 64 & 13.1 \\
(C) VPFB loss without cosine distance & Base & 38.3M & 64 & 11.3 \\
(D) cosine distance $\rightarrow$ inner product & Base & 38.3M & 64 & 9.21 \\
(E) VPFB loss $\rightarrow$ flow matching loss & Base & 38.3M & 64 & 8.90 \\
(F) VPFB loss & Large & 55.7M & 64 & 7.66 \\
(G) VPFB loss & Large & 55.7M & 128 & 6.72 \\
\bottomrule
\end{tabular}%
}
\end{table}

\begin{table}[t]
\caption{Comparison of training-time computational efficiency against EBM baselines.}
\label{tab:training_time}
\centering
\setlength{\aboverulesep}{0pt}
\setlength{\belowrulesep}{1pt}
\resizebox{\columnwidth}{!}{%
\begin{tabular}{lllllll} 
\toprule
\textbf{\multirow{2}{*}{Method}} & \textbf{\multirow{2}{*}{Sampling/Perturbation Approach}}& \textbf{Parameter} & \textbf{Memory} & \textbf{Training} & \textbf{Training} & \textbf{\multirow{2}{*}{FID $\downarrow$}} \\
 &   &  \textbf{Count} &  \textbf{Usage (GB)} &\textbf{Iterations}  & \textbf{Time (hrs)} &  \\
\midrule
VAEBM \cite{VAEBM} & SGLD + Variational Inference + Replay Buffer & 135.9M & \textit{129} & 25K & \textit{414} & 12.2 \\
DRL (\cite{Diffusion_Recovery}) & SGLD + Diffusion & 38.6M & 56 & 240K & 172 & 9.58 \\
CLEL (\cite{CLEL}) & SGLD + Replay Buffer & 30.7M & 10 & 100K & 133 & 8.61 \\
CDRL (\cite{Cooperative_Diffusion_Recovery}) & SGLD + Amortized Inference + Diffusion & 34.8M & 69 & 400K & 144 & 4.31 \\
VPFB-Base (Ours) & Stationary-Enforced OT-FM & 38.3M & 35 & 300K & 48 & 9.33 \\
VPFB-Large (Ours) & Stationary-Enforced OT-FM & 55.7M & 60 & 300K & 112 & 6.72 \\
\bottomrule
\end{tabular}%
}
\end{table}

\begin{table}[t]
\caption{Comparison of inference-time computational efficiency against EBM baselines.}
\label{tab:inference_time}
\centering
\setlength{\aboverulesep}{0pt}
\setlength{\belowrulesep}{1pt}
\resizebox{\columnwidth}{!}{%
\begin{tabular}{lllllll} 
\toprule
\textbf{\multirow{2}{*}{Method}} & \textbf{\multirow{2}{*}{Numerical Approach}}& \textbf{Parameter} & \textbf{Sampling} & \textbf{Inference} & \textbf{Training} & \textbf{\multirow{2}{*}{FID $\downarrow$}} \\
 &   &  \textbf{Count} &  \textbf{Steps} &\textbf{Time (s) }  & \textbf{Latency (ms)} &  \\
\midrule
VAEBM (\cite{VAEBM}) & SGLD & 135.1M & 16 & 21.3 & 13.31 & 12.2 \\
CoopFlow (\cite{CoopFlow}) & SGLD & 45.9M & 30 & 2.5 & 0.833 & 15.8 \\
DRL (\cite{Diffusion_Recovery}) & SGLD & 34.8M & 180 & 23.9 & 1.328 & 9.58 \\
CDRL (\cite{Cooperative_Diffusion_Recovery}) & SGLD & 38.6M & 90 & 12.2 & 1.356 & 4.31 \\
VPFB (Ours) & ODE Solver & 55.7M & 74 & 14.6 & 1.968 & 6.72 \\
\bottomrule
\end{tabular}%
}
\end{table}

\subsection{Computational Efficiency}

Table \ref{tab:training_time} compares the training-time computational efficiency of VPFB against the recent EBM baselines. Here, we included an additional smaller model (VPFB-Base) with fewer parameters but a higher FID score. The training time and GPU memory footprint are measured on a single A100 GPU of 80G memory. Italicized values represent estimates for the VAEBM model based on experiments conducted using smaller batch size, as the model cannot be trained on a single GPU using the prescribed batch sizes. Additionally, Table \ref{tab:inference_time} compares the inference-time computational efficiency of VPFB against the recent EBM baselines. Overall, these results suggest that our method provides improved computational efficiency in both training and inference compared to most strong EBM baselines while maintaining competitive FID scores, particularly for the base model, which has a parameter count similar to the baselines. Nonetheless, there remains a gap in FID performance relative to the state-of-the-art generative models.


\section{Additional Discussions} 
\label{sec:additional_discussions}

In this section, we discuss the strengths and limitations of our proposed VPFB framework in the broader context of energy-based generative modeling. We first highlight the advantages of VPFB over conventional diffusion and flow-based approaches, emphasizing its interpretability, theoretical alignment with Boltzmann energy, and improved sampling efficiency. Next, we examine the critical role of MCMC in achieving Boltzmann-convergent training and sampling. While our method avoids the high cost of long-run MCMC by leveraging structured probability paths, we show that such paths may fail to explore low-density regions in high-dimensional spaces. This limitation can result in mode collapse and degraded sample quality when using deterministic ODE-based sampling. Finally, we analyze the remaining performance gap between VPFB and state-of-the-art EBMs, attributing it to architectural simplicity, trade-offs between convergence and generative sharpness, and the added complexity of modeling marginal rather than conditional distributions. 

\subsection{Advantages over Diffusion and Flow-based Generative Models}

Our proposed energy-parameterized potential flow offers several advantages over conventional diffusion and flow matching models, where vector fields are directly parameterized by neural networks rather than derived as the gradients of scalar-valued energy functions. Specifically, these benefits include:


\textbf{Interpretable Energy Landscape via Marginal Density Modeling} By explicitly parameterizing the vector field as the gradient of a scalar potential energy, our method provides a natural energy-based representation of data dynamics. Such a formulation supports key energy-based modeling tasks, including explicit (marginal) density estimation, composable generation, and OOD detection capabilities not inherently provided by conventional diffusion or flow matching approaches. As shown in Table \ref{tab:auroc}, VPFB demonstrates strong OOD detection performance due to our proposed energy-based formulation.

\textbf{Theoretical Connection to Boltzmann Energy} Our Boltzmann energy formulation in Proposition \ref{thm:proposition_5} rigorously connects the deterministic potential flow to a stationary Boltzmann distribution characterized by the Boltzmann energy $\Phi_B$. This theoretical foundation firmly situates our approach within the energy-based modeling framework, offering theoretical coherence that is lacking in existing diffusion and flow matching models. As a result, it allows our approach to combine the efficiency of flow-based probability paths with the interpretability and rigor of the Boltzmann energy representation. As shown in Figure \ref{fig:density_Boltzmann}, the potential flow converges effectively to the stationary Boltzmann distribution.

\textbf{Optimality of Conservative Vector Field} Our approach learns a purely gradient-based vector field $\nabla_x \Phi$, in contrast to diffusion and flow matching methods, which may incorporate divergence-free components, as noted in \citet{ActionMatching}. By enforcing a conservative energy function through Helmholtz decomposition, our method reduces the dynamical cost associated with these divergence-free components, enabling more efficient particle transport and improving training efficiency, as demonstrated by the comparative benchmark in Table \ref{tab:training_time}.

\textbf{Efficient Deterministic ODE Sampling} By eliminating the reliance on implicit MCMC sampling, our approach reduces computational overhead and avoids common convergence issues encountered in traditional EBMs. Furthermore, our potential flow formulation enables deterministic ODE-based sampling that is generally more stable and efficient for generating high-quality samples with fewer steps than stochastic sampling methods, as demonstrated by the comparative results in Table \ref{tab:inference_time}.

\subsection{Incorporating Langevin Dynamics for Boltzmann-Convergent Sampling}

Conventional EBM training often relies on convergent (long-run) MCMC sampling to thoroughly explore the data space and assign appropriate energy values across the landscape, particularly in low-density regions. This process helps smooth out sharp local minima and mitigates overfitting to high-density areas. In contrast, our framework employs a conditional homotopy ${\rho}(x \mid \bar{x}, t)$ (perturbation kernel) to guide training samples along a structured dynamic probability path. As noted by \citet{Shortrun_MCMC}, such probability paths resemble short-run MCMC behavior, which limits their capacity to explore low-density regions in high-dimensional spaces. Consequently, many of these regions remain unseen during training and are thus poorly modeled in the resulting energy landscape. ODE-based samplers further exacerbate this issue due to their lack of stochasticity, making them more susceptible to becoming trapped in sharp local minima and suffering from poor mixing. Compared to stochastic samplers like Stochastic Gradient Langevin Dynamics (SGLD), deterministic sampling is inherently more sensitive to gaps in energy coverage. This limitation is not unique to our approach - it is a general challenge for diffusion-based and flow-based models that rely on time-dependent Gaussian perturbations to construct their training trajectories.

To achieve proper Boltzmann convergence and reduce mode collapse under deterministic ODE sampling, it is necessary to improve data space coverage beyond what is provided by structured Gaussian perturbations. This could be addressed by incorporating long-run MCMC during training, although doing so incurs the substantial computational overhead characteristic of traditional EBMs. As a potential direction for future work, we propose integrating long-run MCMC sampling within the reverse-diffusion conditional path, following techniques developed by \citet{Diffusion_Recovery},\citet{Cooperative_Diffusion_Recovery}, and \citet{DDAEBM}. However, these methods primarily model the conditional distribution $p(x_{t-1} \mid x_t)$ rather than the marginal data distribution $p(x)$, which remains the focus of our current work. As a result, these conditional EBMs do not establish a direct connection to the Boltzmann distribution $p(x) \propto e^{\Phi(x)}$, which forms the theoretical foundation of marginal energy-based modeling. Adapting such conditional modeling techniques to the marginal EBM setting would require substantial methodological developments. To offset the training cost of long-run MCMC, future work may also explore hybrid strategies involving pre-sampled replay buffers or distillation from a pre-trained convergent EBM.

\subsection{Addressing Performance Gaps: Modeling Trade-offs and Theoretical Challenges}

While our proposed VPFB framework achieves competitive FID performance, there remains a noticeable gap compared to state-of-the-art EBMs. In the following discussion, we analyze the underlying factors contributing to this discrepancy.

\textbf{Difference in Model Architecture} DDAEBM \citep{DDAEBM} utilizes a multi-model architecture comprising three distinct components: (1) a generator model parameterized by the modified U-Net architecture of \citet{diffgan}, (2) an energy model parameterized by the NCSN++ architecture from \citet{NCSNpp}, and (3) a CNN-based encoder. This multi-component architecture enables specialized modules to collaboratively refine each other's behavior through adversarial training, thereby enhancing generative performance. In contrast, our VPFB model adopts a simpler framework design, employing only a single EBM parameterized by the U-Net architecture described in \citet{Dhariwal}. Although this single-component architecture has a comparable parameter count to that of the NCSN++ used in DDAEBM, it may lack the collaborative optimization and mutual refinement advantages that arise from multi-model training setups. Consequently, our VPFB’s FID scores align more closely with those of single-energy or joint-energy EBMs \citep{EnergyorScore,Diffusion_Recovery,CLEL}. We hypothesize that incorporating additional auxiliary model components with adversarial training strategies, as utilized by DDAEBM, could enhance the representational capacity and further sharpen the energy landscape of our VPFB model, potentially leading to improved FID performance. Investigating this hybrid approach, which balances computational efficiency with adversarial training strategy, will be reserved for our future work.

\textbf{Boltzmann stationarity and FID Trade-off} In our previous ablation study, we observed a notable trade-off between Boltzmann stationarity and FID performance. This observation aligns with insights from \citet{NonConvergentMCMC} and \citet{Nijkamp_2020}, which show that non-convergent EBMs outperform convergent EBMs trained with long-run MCMC sampling in image generation. Models such as DDAEBM fall within this class of non-convergent EBMs. We hypothesize that this is due to the inherent tension between accurate equilibrium modeling and sharp generative quality. Specifically, imposing strict Boltzmann convergence tends to smooth out the energy landscape, inadvertently flattening local minima that correspond to meaningful data modes. Although such smoothing enhances theoretical interpretability by faithfully approximating the true equilibrium of the Boltzmann distribution, it compromises the sharpness and detail of generated samples, leading to higher FID scores. To mitigate this limitation, we propose exploring advanced sampling techniques such as the Metropolis-Adjusted Langevin Algorithm (MALA) used in \citet{STIC} or other adaptive short-run samplers in future work. The gradient-informed proposal and acceptance steps of MALA could potentially enhance local mode exploration efficiency without fully abandoning the Boltzmann equilibrium.

\textbf{Conditional vs Marginal EBMs} Another critical contributing factor is the distinction between conditional EBMs as modeled by DDAEBM, and marginal EBMs considered by VPFB. In particular, \cite{DDAEBM} articulate that modeling conditional distributions simplifies the learning task, as these conditional distributions are inherently less multi-modal compared to complex marginal distributions. DDAEBM leverages this insight by decomposing the generation process into discrete diffusion steps, each focusing on simpler, conditional distributions that are easier to model effectively. In contrast, our VPFB explicitly models a global marginal distribution, thus inherently facing greater complexity due to increased modality. Consequently, achieving comparable generative performance poses additional challenges. To address this limitation, future work could explore hierarchical or multi-stage conditional modeling strategies to simplify explicit marginal modeling. Nevertheless, conditional EBMs inherently lack a direct relationship to the marginal Boltzmann distribution, which is fundamental to the theoretical underpinnings of our VPFB framework. Therefore, adapting conditional modeling techniques to fully marginal EBMs would require substantial development.

\newpage
\section{Proofs and Derivations}
\label{sec:proofs_and_derivations}

\subsection{Proof of Proposition \ref{thm:proposition_1}} \label{Appendix:A}
\begin{proof}
Based on the definitions of $q(x)$ and $p(\bar{x} \mid x)$, we can expand their logarithms (ignoring additive constants) as follows:
\begin{align} 
&\log q(x) = -\frac{1}{2 \, \omega^2} \, \frac{1}{2 \, \omega^2} \, \|x\|^2 + \text{(terms independent of }x\text{)} \\
&\log p(\bar{x} \mid x) = -\frac{1}{2 \, \nu^2} \, \|\bar{x} - x\|^2 + \text{(terms independent of }x\text{)}
\end{align}

Substituting these into (\ref{eq:log-homotopy_f}), we obtain:
\begin{align} 
h(x \mid \bar{x}, t) = -\frac{\alpha(t)}{2 \, \omega^2} \, \|x\|^2 - \frac{\beta(t)}{2 \, \nu^2} \, \|\bar{x} - x\|^2
\end{align}

Expanding the squared term:
\begin{align} 
\|\bar{x} - x\|^2 = \|x\|^2 - 2 \, x^T \bar{x} + \|\bar{x}\|^2
\end{align}
and substituting it back into $h(x \mid \bar{x}, t)$:
\begin{align} 
h(x \mid \bar{x}, t) = - \left( \frac{\alpha(t)}{\omega^2} + \frac{\beta(t)}{\nu^2} \right) \|x\|^2 + \frac{\beta(t)}{\nu^2} \, x^T \bar{x}
\end{align}

Recognizing the quadratic form in terms of $x$, we identify that $\rho(x \mid \bar{x}, t)$ is a Gaussian density:
\begin{align} 
{\rho}(x \mid \bar{x}, t) = \mathcal{N} \big(x; \mu(t) \bar{x}, \sigma(t)^2 I \big)
\end{align}
whose mean $\mu(t)$ and variance $\sigma(t)^2$ can be obtained by completing the square.

Define
\begin{align}
A := \frac{\alpha(t)}{\omega^2} + \frac{\beta(t)}{\nu^2}, \qquad B := \frac{\beta(t)}{\nu^2}
\end{align}
Then the exponent becomes
\begin{align}
h(x \mid \bar{x}, t) = -\frac{1}{2} \Bigl[ A \, \|x\|^2 - 2 B \, x^T \bar{x} \Bigr]
\end{align}
and we wish to express this quadratic form as follows
\begin{align}
A \, \|x - \mu(t) \, \bar{x}\|^2 + \text{(terms independent of }x\text{)}
\end{align}
Expanding $A \, \|x - \mu(t) \, \bar{x}\|^2$, we obtain
\begin{align}
A \, \|x - \mu(t) \, \bar{x}\|^2 &= A \, \|x\|^2 - 2 A\, \mu(t)\, x^T \bar{x} + A \, \mu(t)^2 \, \|\bar{x}\|^2
\end{align}
To match the linear term, the mean of the Gaussian is thus given by
\begin{align} \label{eq:mu(t)}
\mu(t) = \frac{B}{A} 
= \frac{\beta(t)/\nu^2}{\alpha(t)/\omega^2 + \beta(t)/\nu^2} 
= \sgmd \Bigg( \log \bigg( \frac{\beta(t)}{\alpha(t)} \, \frac{\omega^2}{\nu^2} \bigg) \Bigg)
\end{align}
where $\sgmd(z) = \frac{1}{1+e^{-z}}$ denotes the standard logistic (sigmoid) function.

By comparing with the standard Gaussian exponent
\begin{align}
-\frac{1}{2 \sigma^2} \, \|x - \mu(t) \, \bar{x}\|^2,
\end{align}
we deduce that the variance is given by
\begin{align}
\sigma(t)^2 = \frac{1}{A} = \frac{1}{\alpha(t)/\omega^2 + \beta(t)/\nu^2}.
\end{align}
Using the expression obtained for $\mu(t)$, the standard deviation can also be written as
\begin{align} 
\sigma(t) = \sqrt{\frac{\nu^2}{\beta(t)} \, \mu(t)}.
\end{align}

\end{proof}

\subsection{Proof of Proposition \ref{thm:proposition_2}} \label{Appendix:B}
\begin{proof}
Differentiating the conditional homotopy ${\rho}(x \mid \bar{x},t)$ in (\ref{eq:unmarginalized_homotopy_h}) with respect to $t$, we have:
\begin{align} \label{eq:homotopy_derivative}
\begin{split}
&\frac{\partial {\rho}(x \mid \bar{x},t)}{\partial t} 
= \frac{1}{\int_{\Omega} e^{h(x \mid \bar{x},t)} \, dx} \, \frac{\partial [e^{h(x \mid \bar{x},t)}]}{\partial t}
- \frac{e^{h(x \mid \bar{x},t)}}{[\int_{\Omega} e^{h(x \mid \bar{x},t)} \, dx]^2} \, \frac{\partial [\int_{\Omega} e^{h(x \mid \bar{x},t)} \, dx]}{\partial t} \\
&= \frac{1}{\int_{\Omega} e^{h(x \mid \bar{x},t)} \, dx} \, \frac{\partial [e^{h(x \mid \bar{x},t)}]}{\partial f} \, \frac{\partial h(x \mid \bar{x},t)}{\partial t}
- \frac{e^{h(x \mid \bar{x},t)}}{[\int_{\Omega} e^{h(x \mid \bar{x},t)} \, dx]^2} \, \int_{\Omega} \frac{\partial [e^{h(x \mid \bar{x},t)}]}{\partial f} \, \frac{\partial h(x \mid \bar{x},t)}{\partial t} \, dx \\
&= \frac{e^{h(x \mid \bar{x},t)}}{\int_{\Omega} e^{h(x \mid \bar{x},t)} \, dx} \, \frac{\partial h(x \mid \bar{x},t)}{\partial t}
- \frac{e^{h(x \mid \bar{x},t)}}{\int_{\Omega} e^{h(x \mid \bar{x},t)} \, dx} \, \int_{\Omega} \frac{e^{h(x \mid \bar{x},t)}}{\int_{\Omega} e^{h(x \mid \bar{x},t)} \, dx} \, \frac{\partial h(x \mid \bar{x},t)}{\partial t} \, dx \\
&= {\rho}(x \mid \bar{x},t) \left( \frac{\partial h(x \mid \bar{x},t)}{\partial t} - \int_{\Omega} {\rho}(x \mid \bar{x},t) \, \frac{\partial h(x \mid \bar{x},t)}{\partial t} \, dx \right) \\
&= - \frac{1}{2} \, {\rho}(x \mid \bar{x},t) \, \Bigg( \frac{d\alpha(t)}{dt} \, \frac{x^T \, x}{\omega^2} + \frac{d\beta(t)}{dt} \, \frac{(x - \bar{x})^T \, (x - \bar{x})}{\nu^2} \\
&\qquad\qquad\qquad\qquad- \int_{\Omega} {\rho}(x \mid \bar{x},t) \, \frac{d\alpha(t)}{dt} \, \frac{x^T \, x}{\omega^2} + \frac{d\beta(t)}{dt} \, \frac{(x - \bar{x})^T \, (x - \bar{x})}{\nu^2} \, dx \Bigg)
\end{split}
\end{align}
where we have applied the quotient rule in the first equation and the chain rule in the second equation.

Subsequently, define
\begin{align}
\begin{split}
\gamma(x, \bar{x}, t) = \frac{d\alpha(t)}{dt} \, \frac{\|x\|^2}{\omega^2} + \frac{d\beta(t)}{dt} \, \frac{\|x - \bar{x}\|^2}{\nu^2}
\end{split}
\end{align}
and using the fact that:
\begin{align} \label{eq:marginalized_homotopy_derivative}
\begin{split}
\frac{\partial \bar{\rho}(x,t)}{\partial t} = \frac{\partial \int_{\Omega} {\rho}(x \mid \bar{x},t) \, {p}_{\text{data}}(\bar{x}) \; d\bar{x}}{\partial t} = \int_{\Omega} \frac{\partial {\rho}(x \mid \bar{x},t)}{\partial t} \, {p}_{\text{data}}(\bar{x}) \; d\bar{x}
\end{split}
\end{align}
we can substitute (\ref{eq:homotopy_derivative}) into (\ref{eq:marginalized_homotopy_derivative}) to obtain:
\begin{align} \label{eq:marginalized_homotopy_derivative_expand}
\begin{split}
\frac{\partial \bar{\rho}(x,t)}{\partial t} = - \int_{\Omega} {p}_{\text{data}}(\bar{x}) \, {\rho}(x \mid \bar{x},t) \, \bigg( \gamma(x, \bar{x}, t) - \int_{\Omega} {\rho}(x \mid \bar{x},t) \, \gamma(x, \bar{x}, t) \, dx \bigg) \; d\bar{x}
\end{split}
\end{align}
Given that both ${\rho}(x \mid \bar{x},t)$ and ${p}_{\text{data}}(\bar{x})$ are normalized (proper) density functions, writing (\ref{eq:marginalized_homotopy_derivative_expand}) in terms of expectations yields the PDE in (\ref{eq:homotopy_PDE}). 

\end{proof}


\subsection{Proof of Proposition \ref{thm:proposition_3}} \label{Appendix:C}
Here, we used the Einstein tensor notation interchangeably with the conventional notation for vector dot product and matrix-vector multiplication in PDE. 
Also, we omit the time index $t$ of $\Phi(x,t)$ in this section for brevity. 

\begin{proof}
Applying forward Euler to the particle flow ODE (\ref{eq:particle_flow_ODE}) using step size $\Delta_t$, we obtain:
\begin{align} 
\begin{split} \label{eq:forward_Euler_rv}
x_{t+\Delta_t} &= \alpha(x_t) = x_t + \Delta_t \, u(x_t) 
\end{split}
\end{align}
where
\begin{align} 
\begin{split} \label{eq:forward_Euler_rv_1}
u(x_t) &= \nabla_{x} \Phi(x_t)
\end{split}
\end{align}
where $x_t$ denotes the discretization of $x(t)$. 
Hereafter, we abbreviate $x_t$, $\alpha(x_t)$, $\nu(x_t)$ as $x$, $\alpha$, $\nu$, respectively.

Assuming that the $\alpha: \Omega \rightarrow \Omega$ is a diffeomorphism (bijective function with differentiable inverse), the push-forward operator $\alpha_{\#}: \mathbb{R} \rightarrow \mathbb{R}$ defines the density transformation ${\rho}^{\Phi}(\alpha,t+\Delta_t):= \alpha_{\#} {\rho}^{\Phi}(x,t)$.
Associated with this change of variable formula is the following density transformation:
\begin{align} 
\begin{split} \label{eq:change_of_variables}
{\rho}^{\Phi}\big(\alpha, t+\Delta_t\big) &= \frac{1}{|D_x \alpha|} \, {\rho}^{\Phi}(x,t)
\end{split}
\end{align}
where $|D_x \alpha|$ denotes the Jacobian determinant of $\alpha$, where the the Jacobian is taken with respect to $x$.

From (\ref{eq:homotopy_PDE}) and (\ref{eq:marginalized_homotopy_derivative_expand}), we obtain:
\begin{align} \label{eq:marginalized_homotopy_derivative_repeat}
\begin{split}
\frac{\partial \log \bar{\rho}(x,t)}{\partial t} &= \frac{1}{\bar{\rho}(x,t)} \, \frac{\partial \bar{\rho}(x,t)}{\partial t} 
= - \, \frac{1}{\bar{\rho}(x,t)} \, \frac{1}{2} \, \mean_{{p}_{\text{data}}(\bar{x})} \Big[{\rho}(x \mid \bar{x}, t) \, \big( \gamma(x, \bar{x}, t) - \bar{\gamma}(x, \bar{x}, t) \big) \Big]
\end{split}
\end{align}

Applying the forward Euler method to (\ref{eq:marginalized_homotopy_derivative_repeat}), we obtain:
\begin{align} 
\begin{split} \label{eq:forward_Euler_homotopy}
\log \bar{\rho}(x, t+\Delta_t) \geq \log \bar{\rho}(x,t) - \frac{\Delta_t}{2} \, \frac{1}{\bar{\rho}(x,t)} \, \mean_{{p}_{\text{data}}(\bar{x})} \Big[{\rho}(x \mid \bar{x}, t) \, \big( \gamma(x, \bar{x}, t) - \bar{\gamma}(x, \bar{x}, t) \big) \Big]
\end{split}
\end{align}

Applying the change-of-variables formula (\ref{eq:change_of_variables}) and then substituting (\ref{eq:forward_Euler_homotopy}) into the KL divergence $\KLD \big[ {\rho}(x, t + \Delta_t) \| \bar{\rho}(x, t + \Delta_t) \big]$ at time $t + \Delta_t$, we obtain:
\begin{align}
\begin{split} \label{eq:min_KLD_problem_expand}
&\KLD\big[ {\rho}^{\Phi}(x, t+\Delta_t) \| \bar{\rho}(x, t+\Delta_t) \big]
= \int_{\Omega} {\rho}^{\Phi}(x,t) \, \log \Bigg( \frac{{\rho}^{\Phi}\big(\alpha, t+\Delta_t\big)}{\bar{\rho}\big(\alpha, t+\Delta_t\big)} \Bigg) \, dx \\
&= \int_{\Omega} {\rho}^{\Phi}(x,t) \, \bigg( \log {\rho}^{\Phi}(x,t) - \log |D_x \alpha| - \log \bar{\rho}\big(\alpha,t\big) \\
&\qquad\qquad\qquad\;\;\;+ \frac{\Delta_t}{2} \, \frac{1}{\bar{\rho}\big(\alpha,t\big)} \, \mean_{{p}_{\text{data}}(\bar{x})} \bigg[ {\rho}(\alpha \mid \bar{x},t\big) \, \Big( \gamma\big(\alpha, \bar{x}, t\big) - \bar{\gamma}\big(\alpha, \bar{x}, t\big) \Big) \bigg] + C \bigg) \, dx \\
\end{split}
\end{align} 

Consider minimizing the KL divergence (\ref{eq:min_KLD_problem_expand}) with respect to ${\alpha}$ as follows:
\begin{align} \label{eq:min_LBKLD_problem}
\begin{split}
\min_{\alpha} \; \KLD (\alpha) &= \min_{\alpha} \; \underbrace{\frac{\Delta_t}{2} \int_{\Omega} {\rho}^{\Phi}(x,t) \, \frac{1}{\bar{\rho}\big(\alpha,t\big)} \, \mean_{{p}_{\text{data}}(\bar{x})} \bigg[ {\rho}\big(\alpha \mid \bar{x},t\big) \, \Big( \gamma\big(\alpha, \bar{x}, t\big) - \bar{\gamma}\big(\alpha, \bar{x}, t\big) \Big) \bigg] \, dx}_{\mathcal{D}^\mathrm{KL}_1(\alpha)} \\
&\qquad\qquad\!\!\!- \underbrace{\int_{\Omega} {\rho}^{\Phi}(x,t) \, \log \bar{\rho}(\alpha,t) \, dx}_{\mathcal{D}^\mathrm{KL}_2(\alpha)} 
\quad- \underbrace{\int_{\Omega} {\rho}^{\Phi}(x,t) \, \log |D_x \alpha| \, dx}_{\mathcal{D}^\mathrm{KL}_3(\alpha)} 
\end{split}
\end{align}
where we have neglected the constant terms that do not depend on ${\alpha}$.

To solve the optimization (\ref{eq:min_LBKLD_problem}), we consider the following optimality condition in the first variation of $\KLD$:
\begin{align} \label{eq:optimality_P1}
\mathcal{I}(\alpha,\nu) = \frac{d}{d\epsilon} \, \KLD\big(\alpha + \epsilon \, \nu\big) \, \bigg|_{\epsilon=0} = 0
\end{align}
This condition must hold for all trial functions $\nu$.

Subsequently, taking the variational derivative of the first functional $\mathcal{D}^\mathrm{KL}_1$ in (\ref{eq:min_LBKLD_problem}), we obtain:
\begin{align} \label{eq:I_1_full}
\begin{split}
&\mathcal{I}^{1}(\alpha,\nu) = \frac{d}{d\epsilon} \, \mathcal{D}^\mathrm{KL}_1(\alpha + \epsilon \nu) \, \bigg|_{\epsilon=0} \\
&= \frac{\Delta_t}{2} \int_{\Omega} {\rho}^{\Phi}(x,t) \, \frac{d}{d\epsilon} \Bigg\{ \frac{1}{\bar{\rho}(\alpha + \epsilon \nu, t)} \, \mean_{{p}_{\text{data}}(\bar{x})} \Big[ \rho(\alpha + \epsilon \nu \mid \bar{x}, t) \, \big( \gamma(\alpha + \epsilon \nu, \bar{x}, t) - \bar{\gamma}(\alpha + \epsilon \nu, \bar{x}, t) \big) \Big] \Bigg\} \, \Bigg|_{\epsilon=0} \, dx \\
&= \frac{\Delta_t}{2} \int_{\Omega} {\rho}^{\Phi}(x,t) \, \frac{\partial}{\partial \alpha} \Bigg\{ \frac{1}{\bar{\rho}(\alpha,t)} \, \mean_{{p}_{\text{data}}(\bar{x})} \Big[ {\rho}(\alpha \mid \bar{x},t) \, \big( \gamma(\alpha, \bar{x}, t) - \bar{\gamma}(\alpha, \bar{x}, t) \big) \Big] \Bigg\} \; \nu \; dx \\
&= \frac{\Delta_t}{2} \int_{\Omega} {\rho}^{\Phi}(x,t) \, D_x \Bigg\{ \frac{1}{\bar{\rho}(\alpha,t)} \, \mean_{{p}_{\text{data}}(\bar{x})} \Big[ {\rho}(\alpha \mid \bar{x},t) \, \big( \gamma(\alpha, \bar{x}, t) - \bar{\gamma}(\alpha, \bar{x}, t) \big) \Big] \Bigg\} \; (D_x \alpha)^{-1} \; \nu \; dx
\end{split}
\end{align}
where the last equation is due to chain rule $\frac{\partial f}{\partial \alpha} = D_x f \, (D_x \alpha)^{-1}$.

Applying the Taylor series expansion to the derivative $\frac{\partial g}{\partial x_i}(\alpha)$ with respect to $x_i$ yields:
\begin{align}
\begin{split} \label{eq:Taylor_series_expansion_2}
\frac{\partial g(\alpha)}{\partial x_i} &= \frac{\partial g(x + \Delta_t u)}{\partial x_i}
= \frac{\partial g(x)}{\partial x_i} + \Delta_t \sum_{j} \frac{\partial^2 g(x)}{\partial x_i \, \partial x_j} \, u_{j} + O(\Delta_t^2) \\
\end{split}
\end{align}

In addition, the inverse of Jacobian $D_x \alpha^{-1}$ can be expanded via the Neuman series to obtain:
\begin{align}
\begin{split} \label{eq:Neuman_series}
D_x \alpha^{-1} &= \big(\, \mathrm{I} + \Delta_t D_x u \,\big)^{-1} 
= \mathrm{I} - \Delta_t D_x u \;+\; O(\Delta_t^2)
\end{split}
\end{align}

Using the Taylor series and Neuman series expansions in (\ref{eq:Taylor_series_expansion_2}) and (\ref{eq:Neuman_series}), we can write (\ref{eq:I_1_full}) in tensor notation, as follows:
\begin{align} \label{eq:I_1_full_expand}
\begin{split}
\mathcal{I}^{1}(\alpha,\nu) 
&= \frac{\Delta_t}{2} \int_{\Omega} {\rho}^{\Phi}(x,t) \, \sum_{i} \, \frac{\partial}{\partial x_i} \bigg\{ \frac{1}{\bar{\rho}(x,t)} \, \mean_{{p}_{\text{data}}(\bar{x})} \Big[ {\rho}(x \mid \bar{x},t) \, \big( \gamma(x, \bar{x}, t) - \bar{\gamma}(x, \bar{x}, t) \big) \Big] \bigg\} \, \nu_{i} \, dx \;+\; O(\Delta_t^2)
\end{split}
\end{align}



Taking the variational derivative of the second functional $\mathcal{D}^\mathrm{KL}_2$ in (\ref{eq:min_LBKLD_problem}) yields:
\begin{align}
\begin{split} \label{eq:I_2_full}
\mathcal{I}^{2}(\alpha,\nu) = \frac{d}{d\epsilon} \, \mathcal{D}^\mathrm{KL}_2(\alpha + \epsilon \nu) \, \bigg|_{\epsilon=0} 
&= \int_{\Omega} {\rho}^{\Phi}(x,t) \, \frac{d}{d\epsilon} \log \bar{\rho}(\alpha + \epsilon \nu) \, \bigg|_{\epsilon=0} \, dx \\
&= \int_{\Omega} {\rho}^{\Phi}(x,t) \, \frac{1}{\bar{\rho}(\alpha, t)} \, \nabla_{x} \bar{\rho}(\alpha, t) \cdot \nu \, dx 
= \int_{\Omega} {\rho}^{\Phi}(x,t) \, \nabla_{x} \log \bar{\rho}(\alpha, t) \cdot \nu \, dx \\
\end{split}
\end{align}
where we have used the derivative identity $d \log g = \frac{1}{g} \, d g$ to obtain the second equation.

Using the Taylor series expansion (\ref{eq:Taylor_series_expansion_2}), we can write (\ref{eq:I_2_full}) in tensor notation, as follows:
\begin{align} \label{eq:I_2_full_expand}
\begin{split}
\mathcal{I}^{2}(\alpha,\nu) &= - \int_{\Omega} {\rho}^{\Phi}(x,t) \, \sum_{i} \Bigg( \frac{\partial \log \bar{\rho}(x,t)}{\partial x_i}
- \Delta_t \sum_{j} \, \frac{\partial^2 \log \bar{\rho}(x,t)}{\partial x_i \, \partial x_j} \, u_{j} \Bigg) \, \nu_{i} \, dx \;+\; O(\Delta_t^2) \\
&= - \int_{\Omega} {\rho}^{\Phi}(x,t) \, \sum_{i} \Bigg( \frac{\partial \log \bar{\rho}(x,t)}{\partial x_i}
- \Delta_t \sum_{j} \, \frac{\partial^2 \log \bar{\rho}(x,t)}{\partial x_i \, \partial x_j} \, u_{j} \Bigg) \, \nu_{i} \, dx \;+\; O(\Delta_t^2)
\end{split}
\end{align}

Similarly, taking the variational derivative of the $\mathcal{D}^\mathrm{KL}_3$ term in (\ref{eq:min_LBKLD_problem}), we obtain:
\begin{align}
\begin{split} \label{eq:I_3_full}
\mathcal{I}^{3}(\alpha,\nu) = \frac{d}{d\epsilon} \, \mathcal{D}^\mathrm{KL}_3(\alpha + \epsilon \nu) \, \bigg|_{\epsilon=0} 
&= \int_{\Omega} {\rho}^{\Phi}(x,t) \, \frac{d}{d\epsilon} \log \big| D (\alpha + \epsilon \nu) \big| \, \bigg|_{\epsilon=0} \, dx \\
&= \int_{\Omega} {\rho}^{\Phi}(x,t) \, \frac{1}{\big|D_x \alpha\big|} \, \frac{d}{d\epsilon} \big| D(\alpha + \epsilon \nu) \big| \, \bigg|_{\epsilon=0} \, dx 
= \int_{\Omega} {\rho}^{\Phi}(x,t) \, \tr \big({D_x \alpha}^{-1} D\nu\big) \, dx
\end{split}
\end{align}
where we have used the following Jacobi's formula:
\begin{align} \label{eq:Jacobi_formula}
\frac{d}{d\epsilon} \big|D (\alpha + \epsilon \nu) \big| \, \bigg|_{\epsilon=0} = \left|D_x \alpha\right| \tr\left(D_x \alpha^{-1}D\nu\right)
\end{align}
to obtain the last equation in (\ref{eq:I_3_full}).

Substituting in (\ref{eq:Neuman_series}) and using the Taylor series expansion (\ref{eq:Taylor_series_expansion_2}), (\ref{eq:I_2_full}) can be written in tensor notation as follows:
\begin{align} \label{eq:I_3_full_expand}
\begin{split} 
\mathcal{I}^{3}(\alpha,\nu) &= \int_{\Omega} \sum_{i} \Bigg( {\rho}^{\Phi}(x,t) \, \frac{\partial \nu_{i}}{\partial x_{i}} - \Delta_t \sum_{j} \, {\rho}^{\Phi}(x,t) \, \frac{\partial u_{j}}{\partial x_{i}} \, \frac{\partial \nu_{i}}{\partial x_{j}} \Bigg) \, dx \,+\, O(\Delta_t^2) \\
&= \int_{\Omega} \sum_{i} \Bigg( \frac{\partial {\rho}^{\Phi}(x,t)}{\partial x_{i}} \, \nu_{i} - \Delta_t \sum_{j} \, \frac{\partial}{\partial x_{j}} \bigg\{ {\rho}^{\Phi}(x,t) \, \frac{\partial u_{j}}{\partial x_{i}} \bigg\} \, \nu_{i} \Bigg) \, dx \,+\, O(\Delta_t^2) \\
&= \int_{\Omega} \sum_{i} \Bigg( \frac{\partial {\rho}^{\Phi}(x,t)}{\partial x_{i}} - \Delta_t \sum_{j} \, \frac{\partial}{\partial x_{j}} \bigg\{ {\rho}^{\Phi}(x,t) \, \frac{\partial u_{j}}{\partial x_{i}} \bigg\} \Bigg) \, \nu_{i} \, dx \,+\, O(\Delta_t^2)
\end{split}
\end{align}
where we have used integration by parts to obtain the second equation.

Taking the limit $\lim \Delta_t \rightarrow 0$, the terms $O(\Delta_t^2)$ that approach zero exponentially vanish.
Subtracting (\ref{eq:I_1_full_expand}) by (\ref{eq:I_2_full_expand}) and (\ref{eq:I_3_full_expand}), then equating to zero, we obtain the first-order optimality condition (\ref{eq:optimality_P1}) as follows:
\begin{align} \label{eq:sum_of_derivatives}
\begin{split}
\int_{\Omega} \bar{\rho}(x,t) \, \sum_{i} \; \sum_{j} &-\, \frac{\partial}{\partial x_{i}} \bigg\{ \frac{1}{\bar{\rho}(x,t)} \, \frac{\partial}{\partial x_{j}} \Big\{ \bar{\rho}(x,t) \, u_{j} \Big\} \bigg\} \\
&+ \frac{1}{2} \, \frac{\partial}{\partial x_i} \bigg\{ \frac{1}{\bar{\rho}(x,t)} \, \mean_{{p}_{\text{data}}(\bar{x})} \Big[ {\rho}(x \mid \bar{x},t) \, \big( \gamma(x, \bar{x}, t) - \bar{\gamma}(x, \bar{x}, t) \big) \Big] \bigg\} \; \nu_{i} \, dx = 0
\end{split}
\end{align}
where we have assumed that ${\rho}^{\Phi}(x,t) \equiv \bar{\rho}(x,t)$ holds and have used the following identities:
\begin{align} \label{eq:used_identities_1}
\begin{split}
\frac{\partial \log \bar{\rho}(x,t)}{\partial x_i} &= \frac{1}{\bar{\rho}(x,t)} \, \frac{\partial \bar{\rho}(x,t)}{\partial x_{i}} \\
\frac{\partial^2 \log \bar{\rho}(x,t)}{\partial x_i \, \partial x_j} &= \frac{\partial}{\partial x_i} \bigg(\frac{1}{\bar{\rho}(x,t)} \frac{\partial \bar{\rho}(x,t)}{\partial x_j}\bigg)
\end{split}
\end{align}

Given that $\nu_{i}$ can take any value, equation (\ref{eq:sum_of_derivatives}) holds (in the weak sense) only if the terms within the round bracket vanish. 
Integrating this term with respect to the $x_i$, we obtain:
\begin{align}
\begin{split} \label{eq:weak_sense_integrated}
\sum_{j} \, \frac{\partial}{\partial x_{j}} \Big\{ \bar{\rho}(x,t) \, u_{j} \Big\} 
= \frac{1}{2} \, \mean_{{p}_{\text{data}}(\bar{x})} \Big[ {\rho}(x \mid \bar{x},t) \, \big( \gamma(x, \bar{x}, t) - \bar{\gamma}(x, \bar{x}, t) \big) \Big] \,+\, \bar{\rho}(x,t) \, C 
\end{split}
\end{align}
which can also be written in vector notation as follows:
\begin{align} \label{eq:weak_solution_PDE_full}
\begin{split} 
& \nabla_{x} \cdot \big( \bar{\rho}(x,t) \, u \big) = \frac{1}{2} \, \mean_{{p}_{\text{data}}(\bar{x})} \Big[ {\rho}(x \mid \bar{x},t) \, \big( \gamma(x, \bar{x}, t) - \bar{\gamma}(x, \bar{x}, t) \big) \Big] \,+\, \bar{\rho}(x,t) \, C
\end{split}
\end{align}

To find the scalar constant $C$, we integrate both sides of (\ref{eq:weak_solution_PDE_full}) to obtain:
\begin{align} \label{eq:integration_by_parts_full}
\begin{split} 
\int_{\Omega} \nabla_{x} \cdot \big( \bar{\rho}(x,t) \, u \big) \, dx 
&= \frac{1}{2} \, \int_{\Omega} \mean_{{p}_{\text{data}}(\bar{x})} \Big[ {\rho}(x \mid \bar{x},t) \, \big( \gamma(x, \bar{x}, t) - \bar{\gamma}(x, \bar{x}, t) \big) \Big] \, dx \,+\, \int_{\Omega} \bar{\rho}(x,t) \, C \, dx \\
&= \frac{1}{2} \, \int_{\Omega} \mean_{{p}_{\text{data}}(\bar{x})} \Big[ {\rho}(x \mid \bar{x},t) \, \big( \gamma(x, \bar{x}, t) - \bar{\gamma}(x, \bar{x}, t) \big) \Big] \, dx \,+\, C
\end{split}
\end{align}

Applying the divergence theorem to the left-hand side of (\ref{eq:integration_by_parts_full}), we obtain:
\begin{align} \label{eq:divergence_theorem}
\begin{split} 
\int_{\Omega} \nabla_{x} \cdot \big( \bar{\rho}(x,t) \, u \big) \, dx 
= \int_{\partial \Omega} \bar{\rho}(x,t) \, u \cdot \hat{n} \, dx
\end{split}
\end{align}
where $\hat{n}$ is the outward unit normal vector to the boundary $\partial \Omega$ of $\Omega$.

Given that $\bar{\rho}(x,t)$ is a normalized (proper) density with compact support (vanishes on the boundary), the term (\ref{eq:divergence_theorem}) becomes zero, leading to $C = 0$.
Substituting this result along with $u(x) = \nabla_{x} \Phi(x)$ into (\ref{eq:weak_solution_PDE_full}), we arrive at the following PDE:
\begin{align} \label{eq:Proposition_1_PDE_general}
\begin{split} 
\nabla_{x} \cdot \big( \bar{\rho}(x,t) \, \nabla_{x} \Phi(x) \big) = \frac{1}{2} \, \mean_{{p}_{\text{data}}(\bar{x})} \Big[ {\rho}(x \mid \bar{x},t) \, \big( \gamma(x, \bar{x}, t) - \bar{\gamma}(x, \bar{x}, t) \big) \Big]
\end{split}
\end{align}

Therefore, assuming that the base case ${\rho}_0(x) \equiv \bar{\rho}_0(x)$ holds and that a solution to (\ref{eq:Proposition_1_PDE_general}) exists at every $t$, the proposition follows by the principle of induction.


\end{proof}

\subsection{Proof of Proposition \ref{thm:proposition_4}} \label{Appendix:D}

To show that the conditional and marginal homotopies satisfy the reverse diffusion process, we first express the forward-time SDE and ODE of \citet{NCSNpp}:
\begin{align} \label{eq:forward_tau_ODE_proof}
\begin{split} 
&dx(\tau) = f(\tau) \, x(\tau) \, d\tau + g(\tau) \, dW(\tau) \\
&\frac{dx(\tau)}{d\tau} = f(\tau) \, x(\tau) - \frac{1}{2} \, g(\tau)^2 \, \nabla_{x} \log p(x,\tau) 
\end{split}
\end{align}
in terms of reverse time $t = 1 - \tau$, via applying the change of variable $dt = -d\tau$ as follows:
\begin{align} \label{eq:forward_time_ODE_proof}
\begin{split} 
&dx(t) = - f(t) \, x(t) \, dt + g(t) \, dW(t) \\
&\frac{dx(t)}{dt} = - f(t) \, x(t) + \frac{1}{2} \, g(t)^2 \, \nabla_{x} \log \bar{\rho}(x,t) 
\end{split}
\end{align}
which gives (\ref{eq:forward_time_SDE}) and (\ref{eq:forward_time_ODE}).

Substituting the marginal score $\nabla_{x} \log \bar{\rho}(x,t)$ with the conditional score:
\begin{align} \label{eq:conditional_score}
\begin{split} 
\nabla_{x} \log {\rho}(x \mid \bar{x},t) = \frac{1}{{\rho}(x \mid \bar{x},t)} \, \nabla_{x} {\rho}(x \mid \bar{x},t) = - \frac{\epsilon}{\sigma(t)}
\end{split}
\end{align}
and applying reparameterization $x(t) = \mu(t) \, \bar{x} + \sigma(t) \, \epsilon$ and (\ref{eq:SDE_drift_diffusion}),
we can write the conditional ODE as follows:
\begin{align} \label{eq:conditional_vector_field_correspondence}
\begin{split} 
\frac{dx(t)}{dt} &= v(x \mid \bar{x},t) \\
&=  - f(t) \, x(t) + \frac{1}{2} \, g(t)^2 \, \nabla_{x} \log {\rho}(x \mid \bar{x},t)  \\
&= - f(t) \, x(t) + \frac{1}{2} \, g(t)^2 \, \frac{\epsilon}{\sigma(t)} \\
&= - f(t) \, x(t) + \sigma(t) \, \Big( \dot{\sigma}(t) + f(t) \, \sigma(t) \Big) \, \frac{\epsilon}{\sigma(t)} \\
&= \frac{\dot{\mu}(t)}{\mu(t)} \, \Big( x(t) - \sigma(t) \, \epsilon \Big) + \dot{\sigma}(t) \, \epsilon \\
&= \dot{\mu}(t) \, \bar{x} + \dot{\sigma}(t) \, \epsilon \\
\end{split}
\end{align}
and thus corresponds to the conditional vector field defined in flow matching \citep{FlowMatching}.

Marginalizing (\ref{eq:conditional_vector_field_correspondence}) with respect to 
\begin{align}
\begin{split} 
p_{\text{data}}(\bar{x} \mid x) = \frac{{\rho}(x \mid \bar{x},t) \, p_{\text{data}}(\bar{x})}{\bar{\rho}(x,t)}
\end{split}
\end{align}
and substituting (\ref{eq:marginal_vector_field}) and applying (\ref{eq:conditional_score}), we obtain
\begin{align} \label{eq:marginal_vector_field_correspondence}
\begin{split} 
v(x,t) 
&= \int_{\Omega} \bigg( - f(t) \, x(t) + \frac{1}{2} \, g(t)^2 \, \nabla_{x} \log {\rho}(x \mid \bar{x},t) \bigg) \, \frac{{\rho}(x \mid \bar{x},t) \, p_{\text{data}}(\bar{x})}{\bar{\rho}(x,t)} \, d\bar{x} \\
&= - f(t) \, x(t) + \frac{1}{2} \, g(t)^2 \, \int_{\Omega} \nabla_{x} \log {\rho}(x \mid \bar{x},t) \, \frac{{\rho}(x \mid \bar{x},t) \, p_{\text{data}}(\bar{x})}{\bar{\rho}(x,t)} \, d\bar{x} \\
&= - f(t) \, x(t) + \frac{1}{2} \, g(t)^2 \, \int_{\Omega} \frac{1}{{\rho}(x \mid \bar{x},t)} \, \frac{{\rho}(x \mid \bar{x},t) \, p_{\text{data}}(\bar{x})}{\bar{\rho}(x,t)} \, \nabla_{x} {\rho}(x \mid \bar{x},t) \, d\bar{x} \\
&= - f(t) \, x(t) + \frac{1}{2} \, g(t)^2 \, \frac{1}{\bar{\rho}(x,t)} \, \nabla_{x} \bar{\rho}(x,t) \\
&= - f(t) \, x(t) + \frac{1}{2} \, g(t)^2 \, \frac{1}{\bar{\rho}(x,t)} \, \nabla_{x} \log \bar{\rho}(x,t)
\end{split}
\end{align}
and thus corresponds to the marginal probability flow ODE \ref{eq:forward_time_ODE}. 

\subsection{Proof of Proposition \ref{thm:proposition_5}} \label{Appendix:E}

\begin{proof} \label{pf:proposition_6}
Based on the result of Proposition \ref{thm:proposition_4} and using (\ref{eq:Kolmogorov_forward}), we can express the homotopy matching problem
\begin{align}
\begin{split} 
&\frac{\partial {\rho}_{\Phi}(x,t)}{dt} = \frac{\partial \bar{\rho}(x,t)}{dt}
\end{split}
\end{align}
equivalently as 
\begin{align}
\begin{split} 
&\nabla_x \cdot \Big( {\rho}_{\Phi} \, \nabla_x {\Phi}(x,t) \Big) = \nabla_x \cdot \bigg( {\rho}_{\Phi} \, \Big( - f(t) \, x(t) + \frac{1}{2} \, g(t)^2 \, \nabla_{x} \log \bar{\rho}(x,t) \Big) \bigg)
\end{split}
\end{align}
Given that this matching holds identically, we have
\begin{align} \label{eq:flow_matching}
\begin{split} 
&\nabla_x {\Phi}(x,t) = - f(t) \, x(t) + \frac{1}{2} \, g(t)^2 \, \nabla_{x} \log \bar{\rho}(x,t) \\
\end{split}
\end{align}

Furthermore, given that both the forward-time ODE and SDE of \citet{NCSNpp} exhibit the same marginal probability density $\bar{\rho}(x,t)$, it is shown that they satisfy the following reverse-time SDE:
\begin{align} \label{eq:reverse_tau_SDE}
\begin{split} 
dx(\tau) = \Big( f(\tau) \, x(\tau) - g(\tau)^2 \, \nabla_{x} \log \bar{\rho}(x,t) \Big) \, d\tau + g(\tau) \, dW(\tau)
\end{split}
\end{align}
which reverses the diffusion process as outlined by \citet{Anderson} and \citet{NCSNpp}.
Applying the change of variable $dt = -d\tau$, this reverse-time SDE can similarly be written in terms of $t = 1 - \tau$ as
\begin{align} \label{eq:reverse_time_SDE}
\begin{split} 
dx(t) = - \Big( f(t) \, x(t) - g(t)^2 \, \nabla_{x} \log \bar{\rho}(x,t) \Big) \, dt + g(t) \, dW(t)
\end{split}
\end{align}
where $dW(t)$ does not change sign, since the Wiener process is invariant under time reversal.

Subsequently, the Fokker-Plank dynamic that governs the time evolution of the marginal density homotopy $\bar{\rho}(x,t)$ is given by
\begin{align}
\begin{split} \label{eq:Fokker_Plank}
\frac{\partial \bar{\rho}(x,t)}{\partial t} 
= - \, \nabla_{x} \cdot \bigg( \bar{\rho}(x,t) \, \Big( - f(t) \, x(t) + g(t)^2 \, \nabla_{x} \log \bar{\rho}(x,t) \Big) \bigg) + \frac{1}{2} \, g(t)^2 \, \Delta_x \, \bar{\rho}(x,t)
\end{split}
\end{align}
where $\Delta_x = \nabla_x \cdot \nabla_x$ denotes the Laplacian. 
By substituting (\ref{eq:flow_matching}) into this Fokker-Plank equation, we then have
\begin{align}
\begin{split} \label{eq:Fokker_Plank_Phi}
\frac{\partial \bar{\rho}(x,t)}{\partial t} 
= - \, \nabla_{x} \cdot \bigg( \bar{\rho}(x,t) \, \Big( 2 \, \nabla_x {\Phi}(x,t) + f(t) \, x(t) \Big) \bigg) + \frac{1}{2} \, g(t)^2 \, \Delta_x \, \bar{\rho}(x,t)
\end{split}
\end{align}

At equilibrium $\frac{\partial \bar{\rho}(x,t)}{\partial t} = 0$, the Fokker-Planck equation admits a unique normalized steady-state solution, given by the Boltzmann distribution:
\begin{align}
\begin{split} \label{eq:Boltzmann_solution}
p_{B}(x) \propto  \exp \Bigg( {\frac{2}{g_{\infty}^2} \, \bigg( 2 \, {\Phi}_{\infty}(x) + \frac{f_{\infty}}{2} \, x(t)^T x(t) \bigg) \Bigg)}
\end{split}
\end{align}
when the potential energy function, the drift coefficient, and the diffusion coefficient reach their time-independent steady states ${\Phi}_{\infty}(x)$, $f_{\infty}$ and $g_{\infty}$ at equilibrium. The Boltzmann distribution can then be written in terms of a coherent Boltzmann energy ${\Phi}_{B}$ considered in EBMs, as follows:
\begin{align}
\begin{split} \label{eq:Boltzmann_EBM}
p_{B}(x) =  \frac{e^{{\Phi}_{B}}}{Z}
\end{split}
\end{align}
where
\begin{align}
\begin{split} \label{eq:Boltzmann_energy_proof}
{\Phi}_{B}(x) = \frac{4 \, {\Phi}_{\infty}(x) + f_{\infty} \, \|x\|^2}{g_{\infty}^2}
\end{split}
\end{align}
\end{proof}
and $Z = {\int_{\Omega} e^{\Phi_{B}(x)} \, dx}$ is the normalizing constant.

\subsection{Proof of Proposition \ref{thm:proposition_6}} \label{Appendix:F}

\begin{proof} \label{pf:proposition_2}
The variational loss function in (\ref{eq:variational_functional}) can be written as follows:
\begin{align} \label{eq:variational_functional_rewritten}
\begin{split}
\mathcal{L}(\Phi, t) = &\ \frac{1}{2} \, \mean_{{\rho}(x \mid \bar{x},t) \, {p}_{\text{data}}(\bar{x})} \Big[ \Phi(x) \, \big( \gamma(x, \bar{x}, t) - \bar{\gamma}(x, \bar{x}, t) \big) \Big] 
+ \frac{1}{2} \, \mean_{\bar{\rho}(x,t)} \, \Big[ \big\| \nabla_{x} \Phi(x) \big\|^{2} \Big]
\end{split}
\end{align}
where we have assumed, without loss of generality, that a normalized energy $\bar{E}_{\theta}(x,t) = 0$. For an unnormalized solution $\Phi(x)$, we can always obtain a normalization by subtracting its mean.

The optimal solution $\Phi$ of the functional (\ref{eq:variational_functional_rewritten}) is given by the first-order optimality condition:
\begin{align} \label{eq:optimality_P2}
\mathcal{I}(\Phi,\Psi) &= \frac{d}{d\epsilon} \, \mathcal{L}(\Phi(x) + \epsilon \Psi(x), t) \, \bigg|_{\epsilon=0} = 0
\end{align}
which must hold for all trial function $\Psi$.

Taking the variational derivative of the particle flow objective (\ref{eq:optimality_P2}) with respect to $\epsilon$, we have:
\begin{align} \label{eq:optimality_P2_full}
\begin{split}
&\mathcal{I}(\Phi,\Psi) = \frac{d}{d\epsilon} \, \mathcal{L}(\Phi + \epsilon \Psi) \, \bigg|_{\epsilon=0} \\
&= \frac{1}{2} \, \int_{\Omega \times \Omega} {p}_{\text{data}}(\bar{x}) \, {\rho}(x \mid \bar{x},t) \, \big( \gamma(x, \bar{x}, t) - \bar{\gamma}(x, \bar{x}, t) \big) \, \frac{d}{d\epsilon} (\Phi + \epsilon \Psi) \; d\bar{x} \, dx \\
&\quad\,+ \frac{1}{2} \, \int_{\Omega} \bar{\rho}(x,t) \, \frac{d}{d\epsilon} \big\| \nabla_{x} (\Phi + \epsilon \Psi) \big\|^2 \, dx
\\
&= \frac{1}{2} \, \int_{\Omega \times \Omega} {p}_{\text{data}}(\bar{x}) \, {\rho}(x \mid \bar{x},t) \, \big( \gamma(x, \bar{x}, t) - \bar{\gamma}(x, \bar{x}, t) \big) \, \Psi \; d\bar{x} \, dx
\;+\; \int_{\Omega} \bar{\rho}(x,t) \, \nabla_{x} \Phi \cdot \nabla_{x} \Psi \, dx
\end{split}
\end{align}

Given that $\Phi \in \mathcal{H}^1_0(\Omega; \rho)$, its value vanishes on the boundary $\partial \Omega$. Therefore, the second summand of the last expression in (\ref{eq:optimality_P2_full}) can be written, via multivariate integration by parts, as
\begin{align} \label{eq:integration_by_parts}
\begin{split}
&\int_{\Omega} \bar{\rho}(x,t) \, \nabla_{x} \Phi \cdot \nabla_{x} \Psi = - \int_{\Omega} \nabla_{x} \cdot \big( \bar{\rho}(x,t) \, \nabla_{x} \Phi \big) \, \Psi \, dx
\end{split}
\end{align}

By substituting (\ref{eq:integration_by_parts}) into (\ref{eq:optimality_P2_full}), we get
\begin{align} \label{eq:optimality_P2_full_simplify}
\begin{split}
\mathcal{I}(\Phi,\Psi)
&= \int_{\Omega} \bigg( \, \frac{1}{2} \, \int_{\Omega} {p}_{\text{data}}(\bar{x}) \, {\rho}(x \mid \bar{x},t) \, \big( \gamma(x, \bar{x}, t) - \bar{\gamma}(x, \bar{x}, t) \big) \; d\bar{x}
\,-\, \int_{\Omega} \nabla_{x} \cdot \big( \bar{\rho}(x,t) \, \nabla_{x} \Phi \big) \bigg) \Psi  \, dx
\end{split}
\end{align}
and equating it to zero, we obtain the weak formulation (\ref{eq:weak_formulation}) of the density-weighted Poisson's equation.

Given that the Poincaré inequality (\ref{eq:Poincare_inequality}) holds,  Theorem 2.2 of \citet{Laugesen} presents a rigorous proof of existence and uniqueness for the solution of the weak formulation (\ref{eq:weak_formulation}), based on the Hilbert-space form of the Riesz representation theorem.
\end{proof}
\newpage
\section{Experimental Details}
\label{sec:experimental_details}

\subsection{Model architecture}
Our network architectures for the autonomous and time-varying VPFB models are based on the WideResNet \citep{wideresnet} and the U-Net \citep{Unet}, respectively. For WideResNet, we include a spectral regularization loss during model training to penalize the spectral norm of the convolutional layer. Also, we apply weight normalization with data-dependent initialization \citep{Salimans} on the convolutional layers to further regularize the model's output. Our WideResNet architecture adopts the model hyperparameters reported by \citet{VAEBM}.
For U-Net, we remove the final scale-by-sigma operation \citep{DDPMpp,NCSNpp} and replace it with the Euclidean norm $\frac{1}{2}\|x - f_{\theta}(x)\|^2$ computed between the input $x(t)$ and the output of the U-Net $f_{\theta}(x)$. Our U-Net architecture adopts the hyperparameters used by \citet{FlowMatching}.  In both the WideResNet and U-Net models, we replace LeakyReLU activations with Gaussian Error Linear Unit (GELU) activations \citep{GELU}, which we find improves training stability and convergence.

\subsection{Training}
We use the Lamb optimizer \citep{LAMB} and a learning rate of $10^{-3}$ for all the experiments. We find that Lamb performs better than Adam over large learning rates. 
We use a batch size of 128 and 64 for training CIFAR-10 and CelebA, respectively.
For all experiments, we set a cutoff time of $t_{\text{max}} = 1 - 10^{-5}$, a terminal time of $t_{\text{end}} = 1$, a decay exponent of $\kappa = 1.5$, and a spectral gap constant of $\eta = 10^{-4}$ during training. Here, the mean and standard deviation scheduling functions $\mu(t) = t$ and $\sigma(t) = 1-t$ follow those defined by the OT-FM path. 
All models are trained on a single NVIDIA A100 (80GB) GPU until the FID scores, computed on 5k samples, no longer show improvement. We observe that the models converge within 300k training iterations.

\subsection{Numerical Solver}
In our experiments, the default ODE solver is the black-box solver from the SciPy library using the RK45 method \citep{RK45}, following the approach of \citet{PFGM}. We allow additional ODE iterations to further refine the samples in regions of high likelihood, which we observe improves the quality of the generated images. This is achieved by extending the time horizon; our experiments indicate that setting the terminal time to $t_\text{end} = 1.575$ yields the best ODE sampling results.


\let\AND\relax
\begin{algorithm}[b]
\caption{VPFB Training}
\label{algo:VPFB_training}
\begin{algorithmic}
\STATE {\bfseries input:} Initial energy model ${\Phi}_{\theta}$, mean and standard deviation scheduling functions $\mu(t)$, $\sigma(t)$, cutoff time $t_{\text{max}}$, terminal time $t_{\text{end}}$, decay exponent $\kappa$, spectral gap constant $\eta$, and batch size $B$.
\REPEAT
\STATE Sample observed data $\bar{x}_i \sim {p}_{\text{data}}(\bar{x})$, $t_i \sim \mathcal{U}(0,t_\text{end})$, and $\epsilon_i \sim \mathcal{N}(0,I)$
\STATE Set $t_i = \min(t_i, t_{\text{max}})$
\STATE Sample $x_i \sim {\rho}(x \mid \bar{x},t_i)$ via reparameterization $x_i = \mu(t_i) \, \bar{x}_i + \sigma(t_i) \, \epsilon_i$
\STATE Compute gradient $\nabla_{x} {\Phi}_{\theta}(x_i, t_i)$ w.r.t. $x_i$ via backpropagation
\STATE Calculate VPFB loss $\frac{1}{B} \sum_{i=1}^{B} \mathcal{L}({\Phi}_{\theta},t_i)$
\STATE Backpropagate and update model parameters ${\theta}$
\UNTIL{FID converges}
\end{algorithmic}
\end{algorithm}

\subsection{Datasets}
We conduct our experiments using the CIFAR-10 \citep{cifar10} and CelebA \citep{celeba} datasets. CIFAR-10 consists of $50,000$ training images and $10,000$ test images at a resolution of $32 \times 32$. The CelebA dataset contains $202,599$ face images, with $162,770$ used for training and $19,962$ for testing. Each image is first cropped to $178 \times 178$ before being resized to $64 \times 64$. During resizing, we enable anti-aliasing by setting the antialias parameter to True. Additionally, we apply random horizontal flipping as a data augmentation technique.


\subsection{Quantitative Evaluation}
We employ the FID and inception scores as quantitative evaluation metrics for assessing the quality of generated samples. For CIFAR-10, the FID is computed between $50,000$ samples and the pre-computed statistics from the training set, following \citet{FID}. For CelebA $64 \times 64$, we adopt the setting from \citet{NCSNv2}, computing the FID between $5,000$ samples and the pre-computed statistics from the test set. For model selection, we follow \citet{NCSNpp}, selecting the checkpoint with the lowest FID score, computed on $2,500$ samples every $10,000$ iterations.

\newpage

\begin{figure}[p]
\centering
\includegraphics[width=0.875\columnwidth]{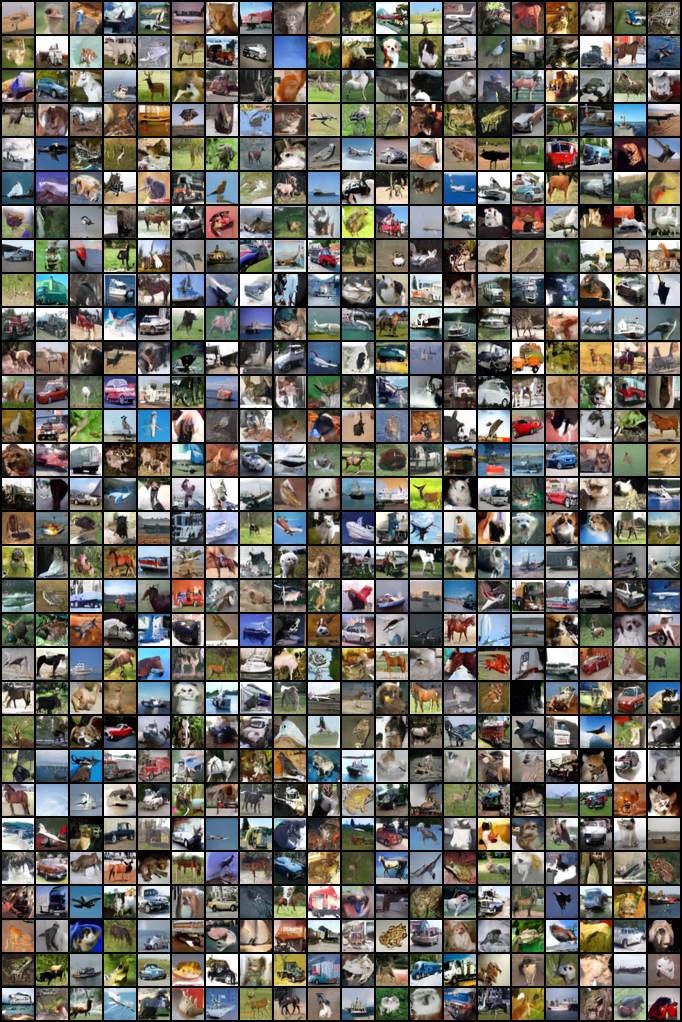}
\caption{Additional uncurated samples on unconditional CIFAR-10 $32 \times 32$.}
\label{fig:cifar10_big}
\end{figure}

\begin{figure}[p]
\centering
\includegraphics[width=0.875\columnwidth]{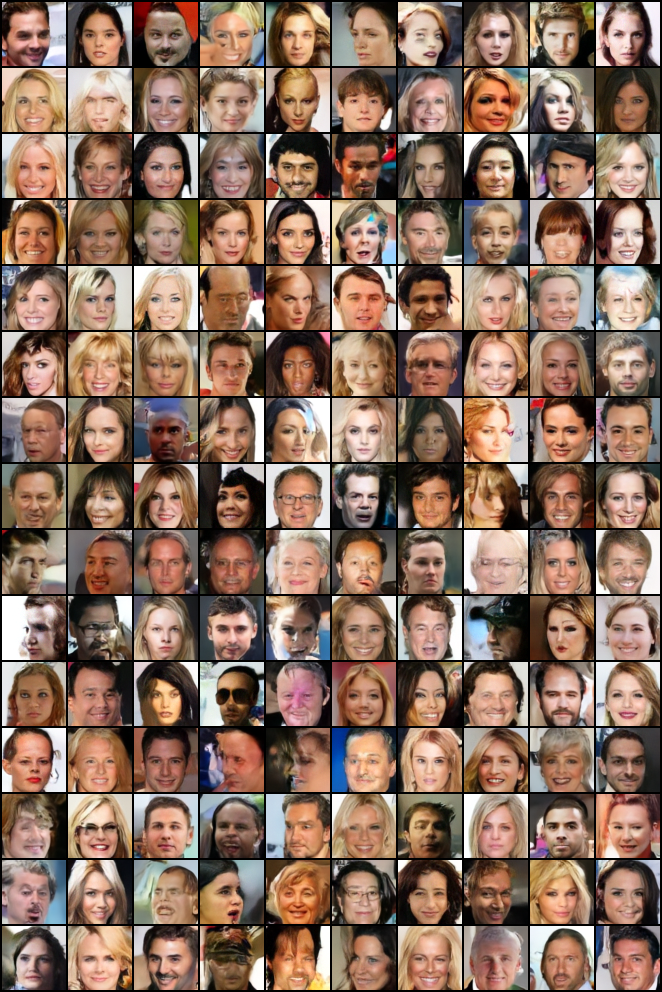}
\caption{Additional uncurated samples on unconditional CelebA $64 \times 64$.}
\label{fig:celeba_big}
\end{figure}

\begin{figure}[p]
\centering
\includegraphics[width=1.0\columnwidth]{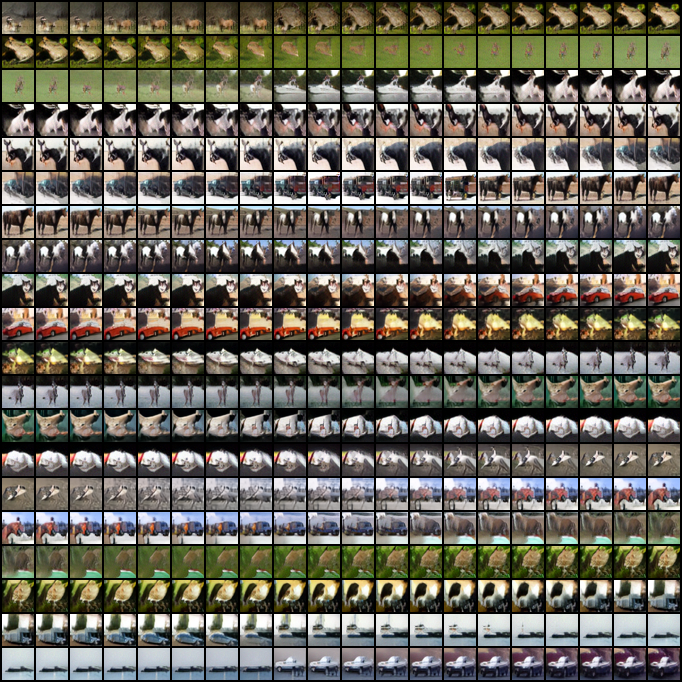}
\caption{Additional interpolation results on unconditional CIFAR-10 $32 \times 32$.}
\label{fig:cifar10_interp_big}
\end{figure}

\begin{figure}[p]
\centering
\includegraphics[width=1.0\columnwidth]{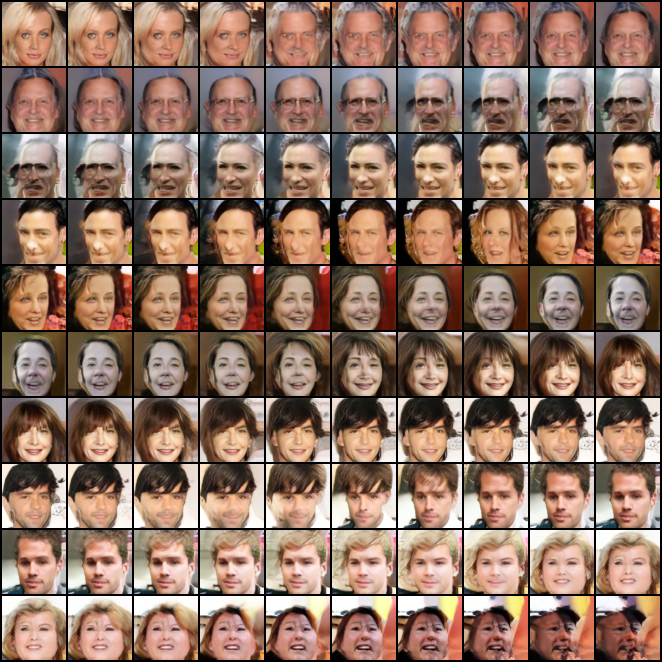}
\caption{Additional interpolation results on unconditional CelebA $64 \times 64$.}
\label{fig:celeba_interp_big}
\end{figure}


\begin{figure}[t]
\centering
\includegraphics[width=1.0\columnwidth]{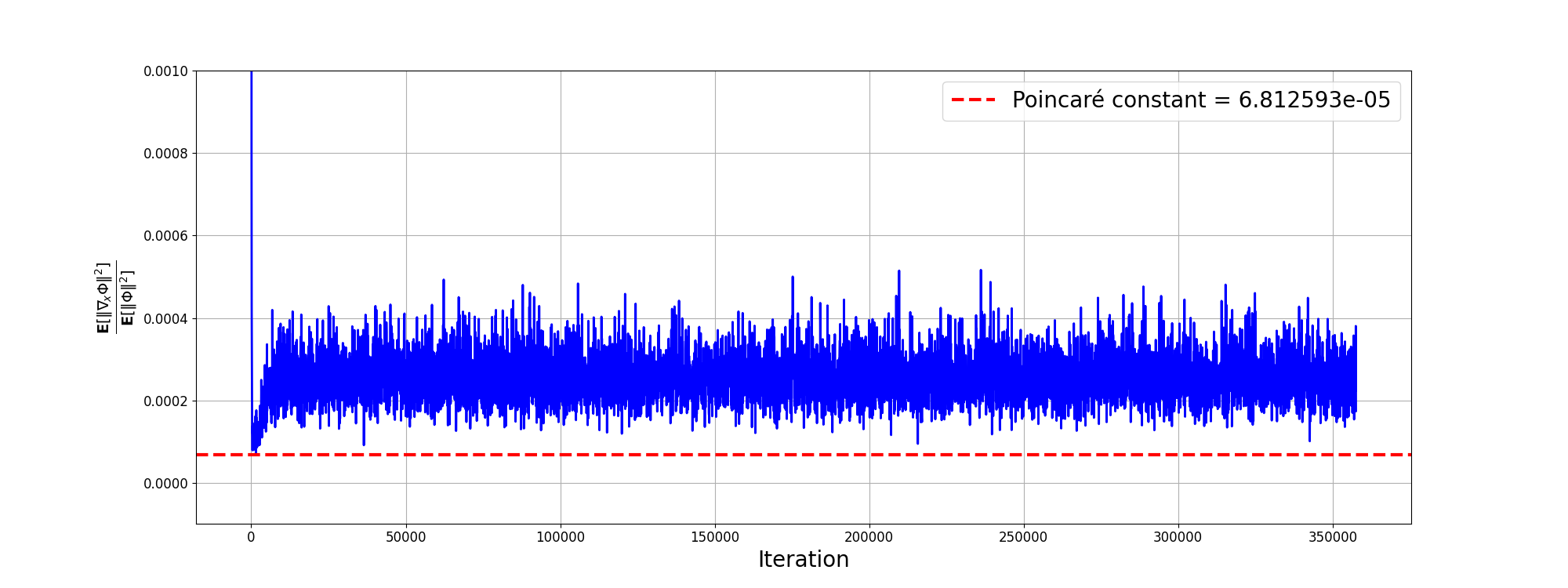}
\caption{Validation of a Poincaré lower bound using the ratio of the gradient norm to the energy norm on CIFAR-10.}
\label{fig:poincare}
\end{figure}

\begin{figure}[t]
\centering
\includegraphics[width=1.0\columnwidth]{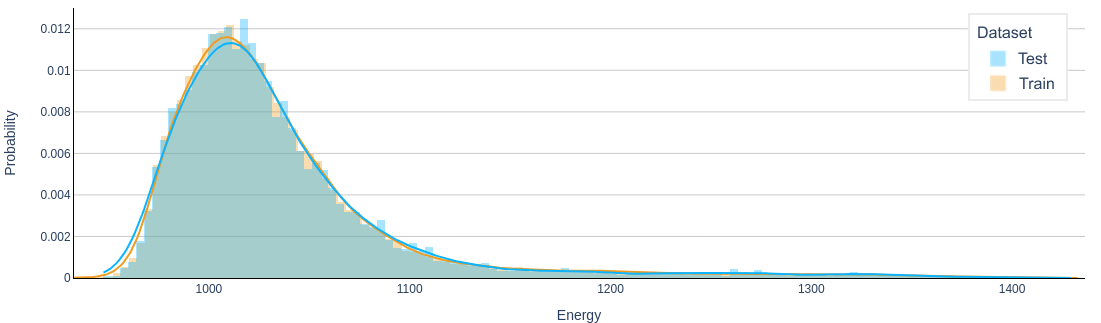}
\caption{Histogram of the energy-parameterized density estimates for the CIFAR-10 training and test datasets.}
\label{fig:histogram_cifar10}
\end{figure}

\begin{figure}[t]
\centering
\includegraphics[width=\columnwidth]{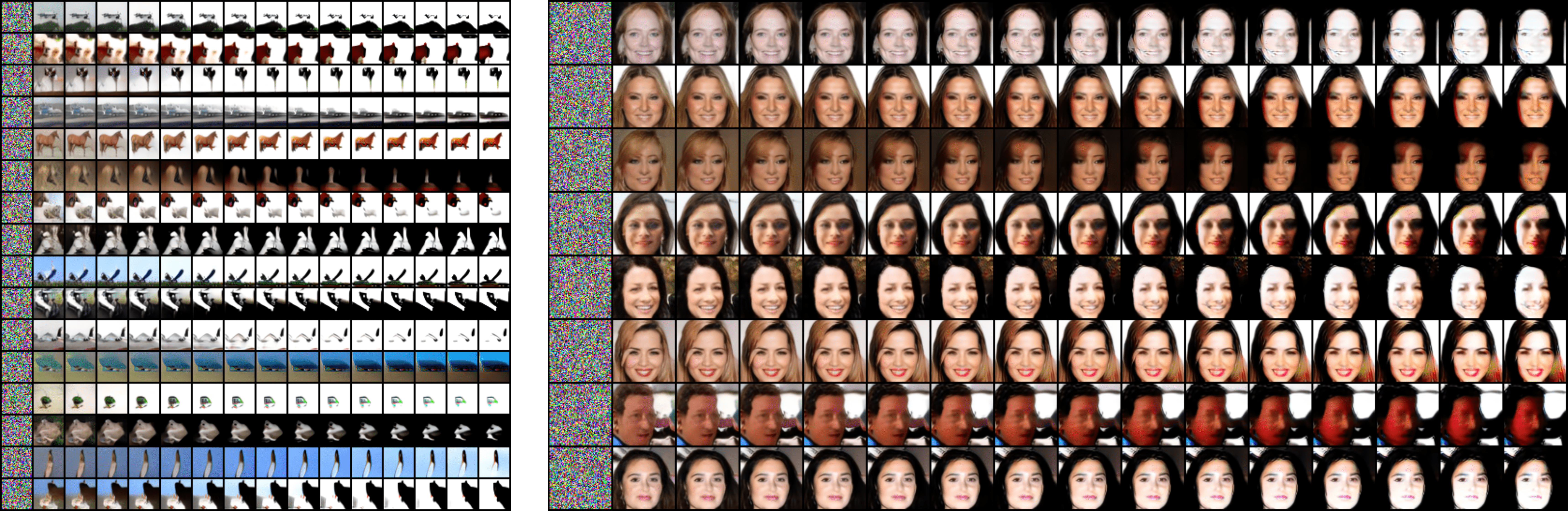}
\caption{Long-run ODE (RK45) sampling using autonomous potential energy $\Phi(x)$ on CIFAR-10 (left) and CelebA (right).}
\label{fig:longrun_ODE_sample_cifar10_celeba}
\end{figure}

\clearpage

\begin{figure}[t]
\centering
\includegraphics[width=1.0\columnwidth]{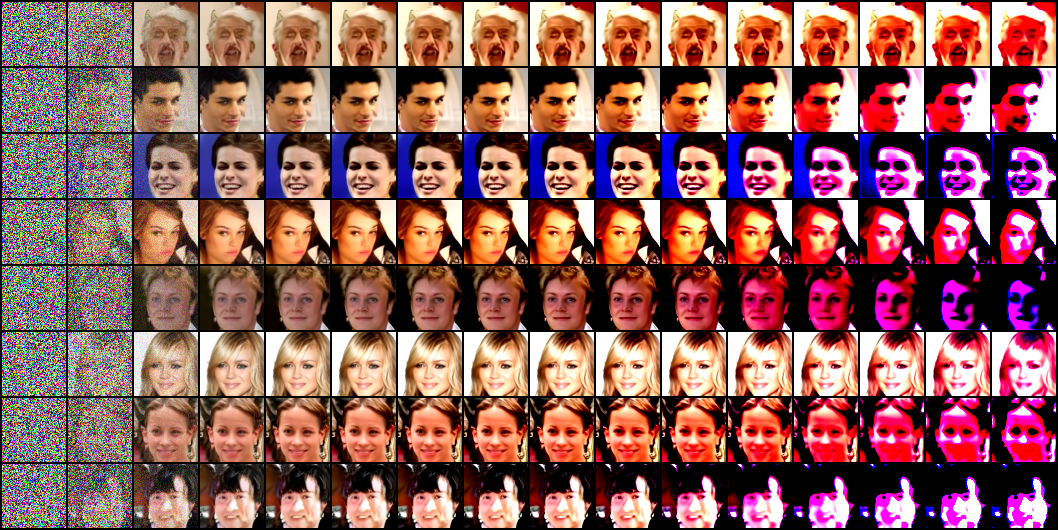}
\caption{Long-run ODE (RK45) sampling using time-varying potential energy $\Phi(x,t)$ on CelebA.}
\label{fig:longrun_ODE_sample_celeba}
\end{figure}

\begin{figure}[t]
\centering
\includegraphics[width=1.0\columnwidth]{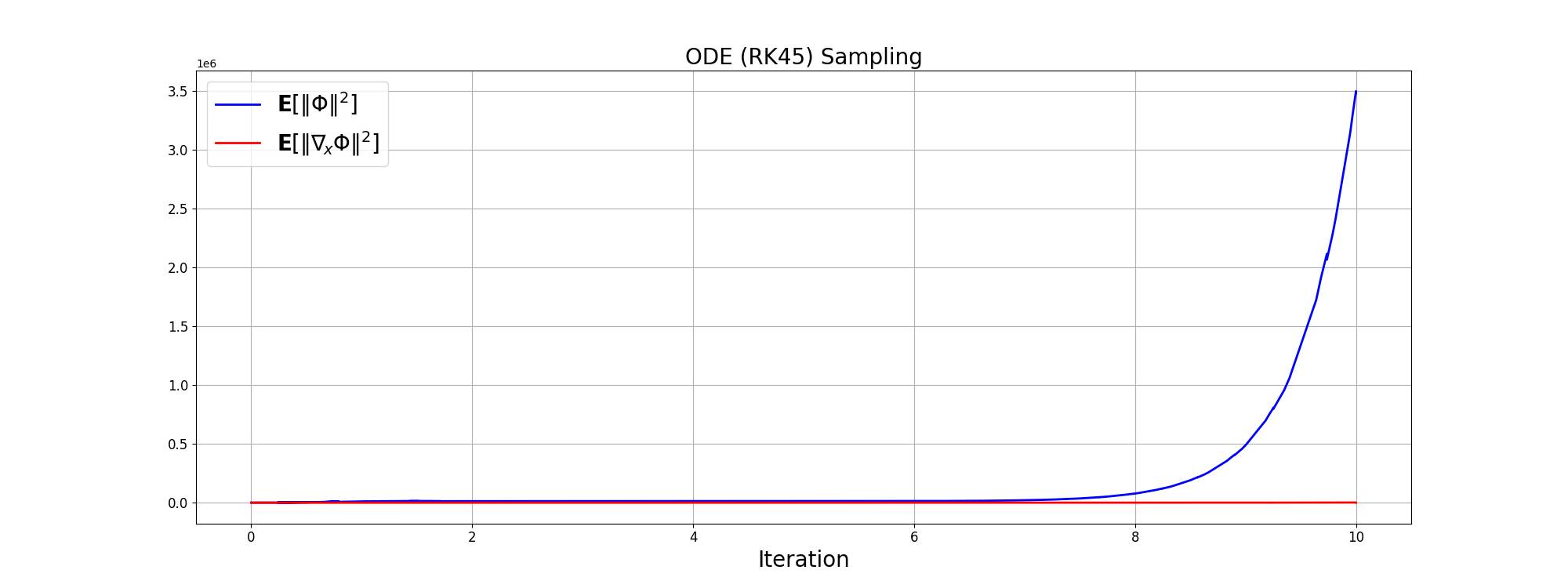}
\caption{Validation of the convergence of gradient norm and energy norm in long-run ODE (RK45) sampling on CelebA.}
\label{fig:longrun_ODE_plot_celeba}
\end{figure}

\begin{figure}[t]
\centering
\includegraphics[width=1.0\columnwidth]{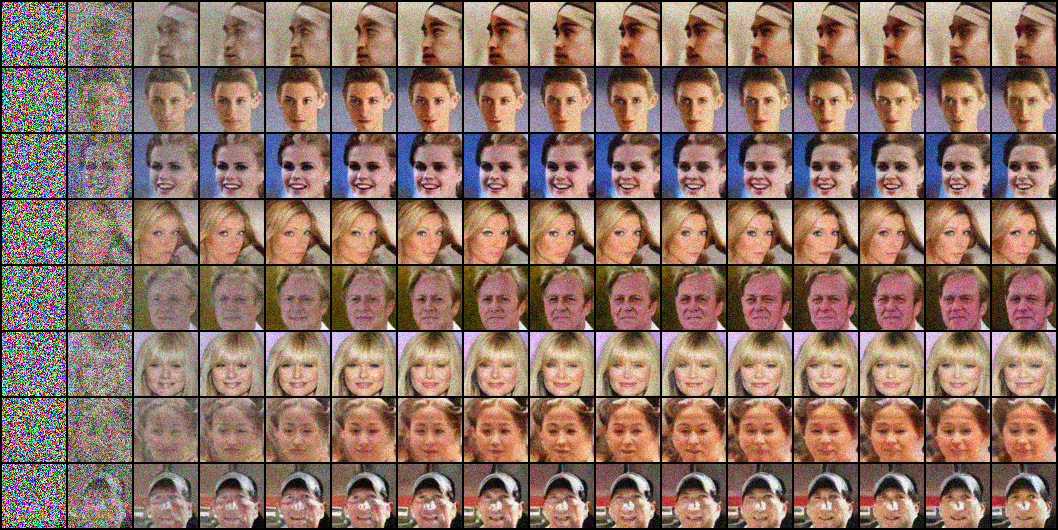}
\caption{Long-run SGLD sampling using the Boltzmann energy with $\lambda=0.35$ on CelebA.}
\label{fig:longrun_SDE_sample_celeba}
\end{figure}

\begin{figure}[t]
\centering
\includegraphics[width=1.0\columnwidth]{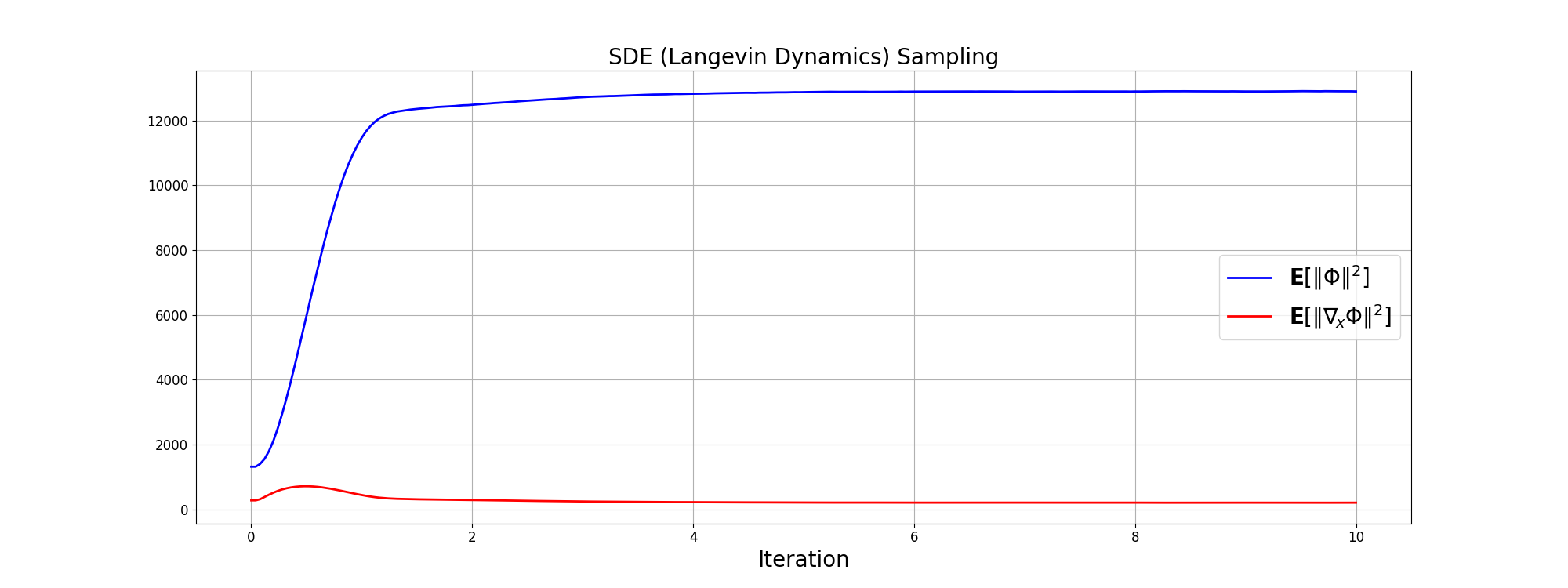}
\caption{Validation of the convergence of the gradient norm and the energy norm in long-run SGLD sampling on CelebA.}
\label{fig:longrun_SDE_plot_celeba}
\end{figure}

\begin{figure}[t]
\centering
\includegraphics[width=0.75\columnwidth]{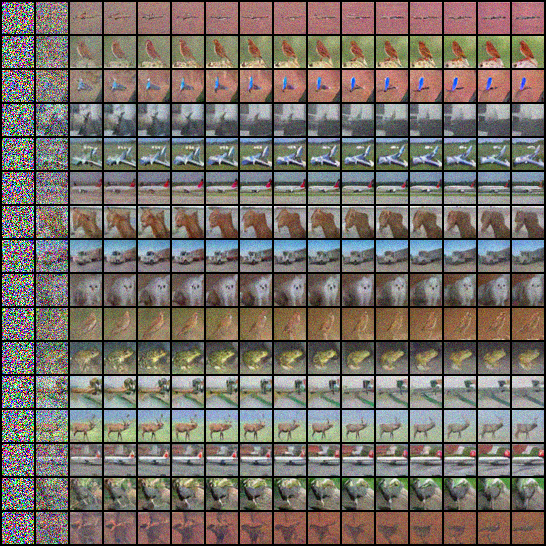}
\caption{Long-run SGLD sampling using the Boltzmann energy with $\lambda=0.35$ on CIFAR-10.}
\label{fig:longrun_SDE_sample_cifar10}
\end{figure}

\begin{figure}[t]
\centering
\includegraphics[width=1.0\columnwidth]{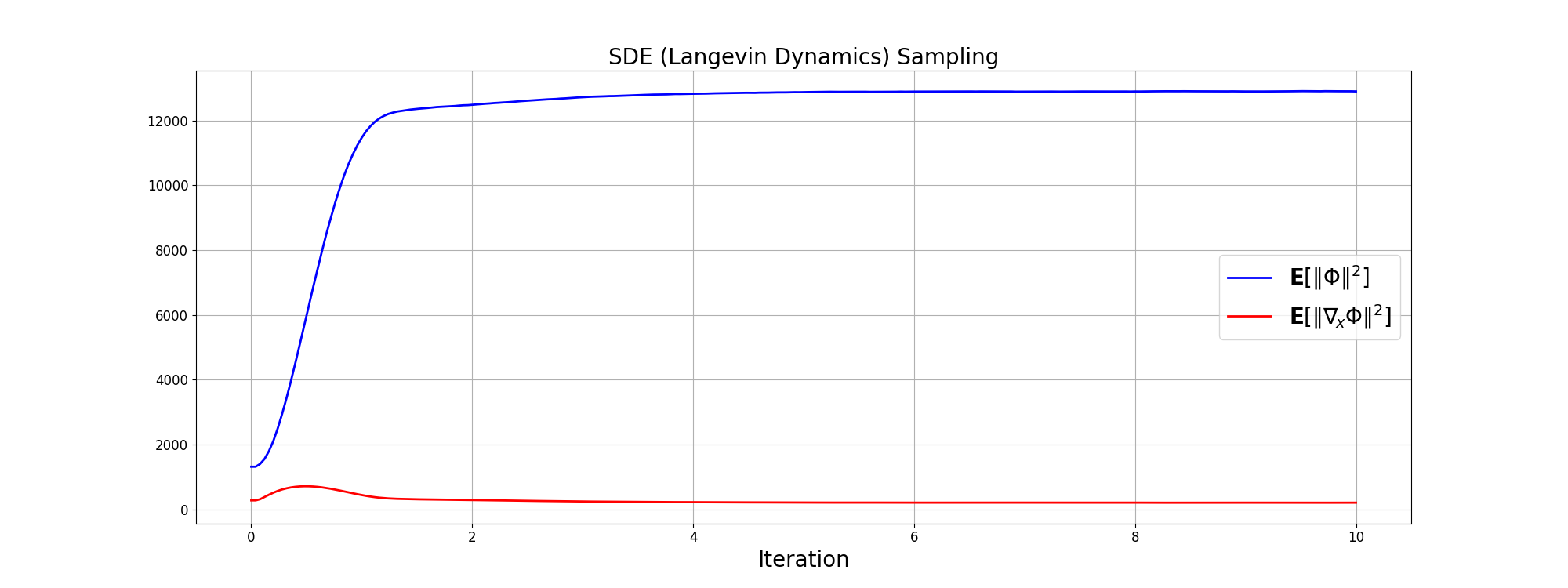}
\caption{Validation of the convergence of the gradient norm and the energy norm in long-run SGLD sampling on CIFAR-10.}
\label{fig:longrun_SDE_plot_cifar10}
\end{figure}

\end{document}